\documentclass[10pt,journal]{IEEEtran}
\usepackage{amsmath,amsfonts}
\usepackage{array}
\usepackage{graphicx}
\usepackage{fancyhdr}

\usepackage[linesnumbered,ruled,vlined]{algorithm2e}
\SetAlCapNameFnt{\small}
\SetAlCapFnt{\small}
\usepackage[dvipsnames]{xcolor}
\usepackage{caption}
\usepackage{subcaption}
\usepackage{floatrow}
\usepackage{amsmath}
\usepackage{amssymb}
\usepackage{array}
\usepackage{tabularx}
\usepackage{tikz}
\usetikzlibrary{positioning, arrows.meta, shadows, fit, calc, intersections}
\usepackage[backref=page]{hyperref}

\usepackage{array}
\usepackage{tabularx}
\newcolumntype{b}{X}
\newcolumntype{s}{>{\hsize=.4\hsize}X}
\newcolumntype{m}{>{\hsize=.7\hsize}X}
\newcolumntype{t}{>{\hsize=.3\hsize}X}
\newcolumntype{v}{>{\hsize=.2\hsize}X}
\newcolumntype{q}{>{\hsize=.1\hsize}X}

\usepackage{threeparttable}

\fancypagestyle{firstpagestyle}
{
  \fancyhf{}
  \fancyhead[L]{\footnotesize{%
    \parbox{\textwidth}{
    \centering
          This paper has been accepted for publication in \emph{IEEE Transactions on Robotics}. \\ DOI: \href{https://doi.org/10.1109/TRO.2023.3326350}{10.1109/TRO.2023.3326350} 
        }}
    }

  \fancyfoot[L]{\footnotesize{%
    \fbox{%
        \parbox{\dimexpr\linewidth-2\fboxsep-2\fboxrule}{%
        \textcopyright 2023 IEEE.  Personal use of this material is permitted.  Permission from IEEE must be obtained for all other uses, in any current or future media, including reprinting/republishing this material for advertising or promotional purposes, creating new collective works, for resale or redistribution to servers or lists, or reuse of any copyrighted component of this work in other works.
  }}}}

}

\begin{document}
\title{Adaptive Asynchronous Control Using Meta-learned Neural Ordinary Differential Equations}
\author{Achkan Salehi$^\dagger$, Steffen Rühl$^{\ddagger}$ and Stephane Doncieux$^\dagger$
\thanks{$\dagger$ Achkan Salehi and Stephane Doncieux are with the ISIR, Sorbonne University, CNRS, Paris, France (email: achkan.salehi@sorbonne-universite.fr, stephane.doncieux@sorbonne-universite.fr).}
\thanks{$\ddagger$ Steffen Rühl is with the perception team at Magazino GmbH (email: ruehl@magazino.eu).}
}



\maketitle

\thispagestyle{firstpagestyle}
\begin{abstract}
  Model-based Reinforcement Learning and Control have demonstrated great potential in various sequential decision making problem domains, including in robotics settings. However, real-world robotics systems often present challenges that limit the applicability of those methods. In particular, we note two problems that jointly happen in many industrial systems: 1) Irregular/asynchronous observations and actions and 2) Dramatic changes in environment dynamics from an episode to another (\textit{e.g.} varying payload inertial properties). We propose a general framework that overcomes those difficulties by meta-learning adaptive dynamics models for continuous-time prediction and control. \color{black} The proposed approach is task-agnostic and can be adapted to new tasks in a straight-forward manner \color{black}. We present evaluations in two different robot simulations and on a real industrial robot.
\end{abstract}

\begin{IEEEkeywords}
  Model Learning for Control, Robust/Adaptive Control of Robotic Systems, Learning and Adaptive Systems, Industrial Robots
\end{IEEEkeywords}

\begin{small}
\section{Introduction}
  \label{sec_intro}

  Machine Learning methods have increasingly been studied and applied for decision making and control in robotic systems during the past few years. Model-free Reinforcement Learning (RL) methods \cite{levine2018learning,openai2019solving,kumar2021rma} have shown great success in various robotics settings, and model-based Reinforcement Learning and Planning \cite{deisenroth2011pilco, pinneri2021sample} have garnered considerable attention due to their increased data-efficiency. More recently, population based methods such as model-free and model-based Quality-Diversity (QD) algorithms \cite{lim2022dynamics, kim2021exploration} have also shown promise in contexts where behavioral diversity is essential.

 \begin{figure}[ht!]
  \centering
  \captionsetup[subfigure]{justification=centering}
      \includegraphics[width=76mm,trim={0cm 0 0cm 0.0cm},clip]{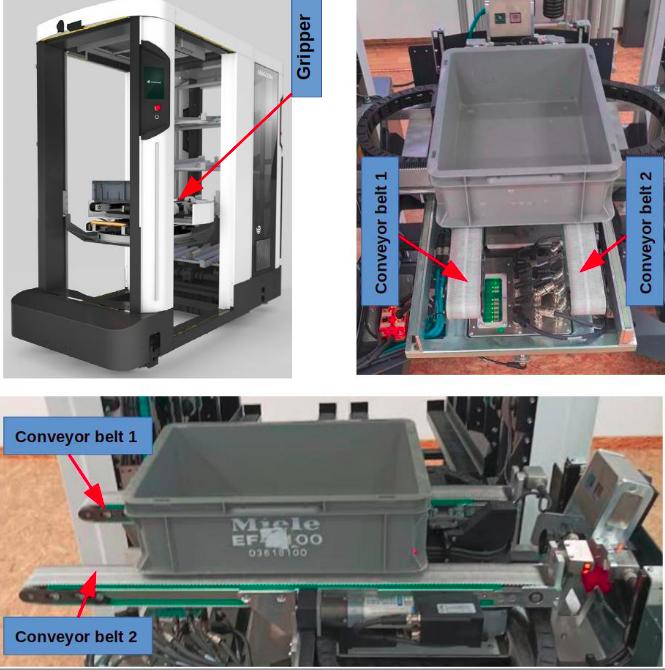}
      \caption{\textbf{(Top left)} The SOTO2 robot manufactured by Magazino GmbH, which autonomously navigates in storage facilities such as warehouses and uses a gripper essentially composed of two independently controlled conveyor belts to manipulate boxes. \textbf{(Top right and bottom)} Different views of the gripper, the control of which requires addressing the two problems mentioned in the introduction: first, commands and observations are sent and received in an irregular/asynchronous manner. Second, the distributions of mass of the boxes that are manipulated can dramatically vary from an episode to the next.}
   \label{intro_figure}
\end{figure}

  However, the vast majority of those methods have been applied in contexts where the difficulties and constraints present in real-world systems are partially relaxed. In particular:

  \begin{enumerate}
    \item {In a large number of systems, observations and actions are irregular/asynchronous. This contrasts with the assumptions made by the majority of works in RL/QD, in which actions are applied at a given state and observations are received at regular intervals.}
    \item{The dynamics of the environments in which real-world systems operate are often subject to dramatic discontinuous changes between episodes. This can be for example because of the change in the geometry or mass distribution of a payload, or an unexpected malfunction of some component.}
  \end{enumerate}

  An example in which those difficulties can be encountered is the industrial SOTO2 robot (figure \ref{intro_figure}), which is manufactured by Magazino GmbH\footnote{https://www.magazino.eu} and deployed in production or assembly lines in order to autonomously manipulate and transport payloads of varying dimensions and mass distributions. In particular, its gripper, which leverages two independent conveyor belts (figure \ref{intro_figure} top right and bottom) receives commands irregularly. Those commands are not synchronized with the \textemdash potentially also irregular \textemdash observations that are provided by its different sensors. Furthermore, a controller aiming to achieve non-trivial goals using those conveyor belts should be able to adapt to the inertial properties of previously unseen payloads.

  \color{black}In addition to adapting to changes in environment dynamics, a desirable property for autonomous robots is the ability to adapt to new, potentially unseen tasks. For example, given the same payload, the SOTO2 might be tasked to rotate it by some desired angle or instead displace it according to a succession of affine transformations or splines with constraints on velocity/accelerations. Similarly, a mobile robot tasked with navigating to specific goals might in the future be required to explore/search its environment instead. Our aim in this paper is to propose a solution that jointly handles the problems of asynchronous/irregular inputs and changes in environment dynamics, while retaining flexibility with respect to downstream tasks. \color{black}

  \color{black}
  Current works in RL/QD and Model-Predictive Control (MPC) in general do not jointly address all of the aforementioned aspects. For example, many RL/QD systems that incorporate meta-learning \cite{finn2017model, openai2019solving, nagabandi2018learning, team2023human, salehi2022few, belkhale2021model} or system identification \cite{kumar2021rma} are able to adapt to unseen environment and/or tasks with minimal adaptation. However, they do not address the problem of irregular actions and observations. On the other hand, RL methods that are designed to handle those issues and/or operate in continuous-time scenarios \cite{yildiz2021continuous, du2020model} are applied to environments that do not change significantly between episodes. Similarly, the incorporation of Neural Ordinary Differential Equations (N-ODES) by a growing body of MPC-based methods \cite{chee2022knode, jiahao2023online,richards2022control} makes them able to process irregular/asynchronous inputs in a natural manner. However, they often focus on accuracy in a single environment (\textit{e.g.} \cite{chee2022knode}), and recent efforts towards adaptive MPC with N-ODEs \cite{jiahao2023online, richards2022control} either adapt to novel environments while gradually forgetting previous ones \cite{jiahao2023online} or learn controllers and models that are specific to trajectory-tracking objectives \cite{richards2022control} which results in reduced flexibility in terms of adaptation to novel tasks, such as exploration.

  In this paper, we propose a general framework, dubbed \textbf{\textit{ACUMEN}}, for \textbf{A}daptive \textbf{C}ontrol in Asynchrono\textbf{U}s proble\textbf{M} s\textbf{E}tti\textbf{N}gs, which is capable of handling irregular/asynchronous observations and actions and changes in environment dynamics in a \textit{task-agnostic} manner. It is based on incorporating meta-learned Neural Ordinary Different Equations (N-ODEs) \cite{chen2018neural} that model environment dynamics into a sampling-based MPC pipeline.
  \color{black}

  We perform experiments in two different robot simulations as well as on the real SOTO2 robot. The first simulation, which we have developed for this work, is a light-weight pybullet \cite{coumans2021} environment that provides a simplified model of the gripper from the SOTO2 robot. The second set of experiments is based on the Gazebo Turtlebot3 simulation \cite{gazeboturtle} which relies on ROS\footnote{The Robot Operating System\cite{rossoft}.} for asynchronous communications with the controller. Finally, our experiments on the real SOTO2 robot demonstrate the feasibility of using the proposed neural-ODE based model predictive approach in real-world robotic systems. \color{black} However, those live experiments do not include meta-learning due to limited access to the robot. Instead, meta-learning results are provided based on off-line data from robots that are deployed at various customer sites \color{black}.

  The paper is structured as follows. The following section (\S\ref{formulation_sec}) is dedicated to problem formalization and notations. We position our work with respect to the literature in \S\ref{sec_related} and describe the proposed method in \S\ref{sec_acumen}. Experimental results in simulated environments and on the real SOTO2 robot are respectively presented in \S\ref{sec_experim_simu} and \S\ref{sec_experim_soto}. Some advantages of the proposed system, its limitations and future directions are discussed in section \S\ref{sec_discussion}. Closing remarks follow in \S\ref{sec_concl}.

\section{Problem formulation and Notations}
  \label{formulation_sec}

  We consider robotic systems with irregular/asynchronous actions and observations that operate in episodic fashion, and where the unknown dynamic system governing state transitions is sampled from an unknown distribution for each new episode. A known reward function, defining the task, can also be sampled for each episode. We formalize those problems below.\\

  \noindent \textbf{Irregular/asynchronous actions and observations.} In most of the RL literature, actions and observations are regular and synchronized: one performs an action $a_t$ in some state $z_t$, and receives an observation $\omega_{t+1}$. In contrast, we are interested in the general case in which observations and actions are given as $\omega_{t_1}, ..., \omega_{t_k}$ and $a_{t'_1}, ..., a_{t'_l}$ with $l\neq k$, $t'_i\neq t_i$. Typically, there might be many actions between two observations, and vice-versa. 

We frame such problems as Continuous Time, Partially Observable Markov Decision Processes (CTPOMDP) in which the observation function is time-dependent. More formally, we consider the tuple $\mathcal{T}\triangleq <Z,\mathcal{A},\mathcal{F},R,\Omega, \Phi,\gamma, T>$ where 
  $Z \subset \mathbb{R}^N$ and $\mathcal{A} \subset \mathbb{R}^M$ respectively denote the state and action spaces. Let us denote $u: \mathbb{R} \rightarrow \mathcal{A}$ the function defining the actions to be applied at time $t$. Then $\mathcal{F}$ defines the evolution of the dynamic system:
  
  \begin{equation}
    z(t_0+\Delta t)=z(t_0)+\int_{t_0}^{\Delta t} \mathcal{F}(z(t),t,u(t))dt.
  \end{equation}

  The possible time intervals over which the system operates are noted $T$ and the function $\Phi: Z \times T \times \Omega \rightarrow [0,1]$ defines the probability $p(\omega| z, t)$ of an element $\omega$ of the observation space $\Omega$ given a state-time pair. In what follows, $a_t$ will denote an action applied at time $t$. Likewise, $\omega_t$ will indicate an observation made at time $t$. The function $R: Z \times \mathcal{A} \rightarrow \mathbb{R}$ is a dense reward function and $\gamma\in [0,1]$ is a discount factor. During an episode associated to the tuple $\mathcal{T}$, our aim is to find $u(t)$ that maximizes the discounted cumulative reward $\int_0^{\infty} \gamma R(z(t), u(t))dt$. \\

  \noindent \textbf{Changes in environment dynamics and tasks.} We consider the case in which environment dynamics remain stationary through the duration of an episode, but dramatically and discontinuously change in between different episodes. More formally, each environment $\mathcal{T}$ is sampled from a distribution of the form

  \begin{equation}
    P(\mathcal{T})=P(<Z,\mathcal{A},\mathcal{F},R,\Omega, \Phi,\gamma, T>)\triangleq P(\mathcal{F},\color{black}R\color{black}).
  \end{equation}

  In other words, two episodes associated to two distinct environments $\mathcal{T}_i, \mathcal{T}_j$ only differ in $\mathcal{F}_i, R_i$ and $\mathcal{F}_j, R_j$, while the rest remains unchanged. Our objective is to find an adaptive control process that leverages previous learning experience to adapt to new, unseen environments from $P(\mathcal{T})$, \color{black} in a manner that is robust to changes in $R$\color{black}.

\section{Related Work}
\label{sec_related}

\color{black} The proposed solution is made of: 1) a sampling based MPC module 2) A dynamics modelling component and 3) a meta-learning framework. In this section, we first discuss positioning w.r.t each of those building blocks separately (\S\ref{sec_related_mb},\ref{sec_related_irreg},\ref{sec_related_meta}). Then, in section \S\ref{sec_related_system}, we discuss related solutions from the RL/Control literatures that combine similar modules to solve problems related to adaptivity and irregularity in the model-based setting. \color{black}

  \subsection{Model-based Reinforcement Learning and Control.} 
  \label{sec_related_mb}

  Model-based RL (MBRL) \cite{moerland2020model} with learned models has been demonstrated to be superior in terms of data-efficiency compared to model-free RL (MFRL), particularly in low-data regimes \cite{kaiser2019model,nagabandi2018neural,deisenroth2011pilco}. This is particularly important on robotic systems, where the large number of interactions required by model-free methods are often impractical. While it has been observed that MBRL can suffer from lower asymptotic performance compared to MFRL \cite{nagabandi2018neural, kaiser2019model}, recent works \cite{chua2018deep} demonstrate that this performance gap can be closed by accounting for model uncertainty, which results in better exploration. 

  \color{black} Closely related to MBRL are Model Predictive Control (MPC) methods, which consist in finding a sequence of controls that optimize a trajectory cost function, given a dynamics model traditionally derived from first principles. Considering such models as prior knowledge, some modern methods leverage machine learning techniques to construct more complex environment models that account for uncertainty \cite{chee2022knode} or inaccuracies \cite{saviolo2022physics}, while others learn the dynamics model from scratch \cite{spielberg2021neural}. We note that the majority of MPC methods minimize a cost function that depends on a target state-space trajectory. However, not all tasks can be specified by a desired path in a straight-forward manner, and it can be preferable \textemdash from both feasibility and generalizability perspectives\textemdash to learn how to infer such trajectories without explicit planning. This is for example the case for tasks that require exploration or precise interactions with the environment. In such cases, the flexibility of RL methods makes them better candidates. \color{black}

\color{black} A popular approach that lies at the frontier between RL and MPC is to sample action-state trajectories from a learned model, and score each of them according to a reward that is (partially) dependent on the predicted states\color{black}. The $N$ first actions from the best trajectory are then applied on the real system. An example of such an action selection scheme is the Cross Entropy Method (CEM) \cite{rubinstein1999cross, pinneri2021sample}. As a population-based algorithm, the model-based CEM is robust to model inaccuracies, and has been shown to produce results that are on-par with MFRL \cite{chua2018deep, hafner2019learning}. Furthermore, CEM, unlike most of the work based on learned policies, is not tied to a particular reward function: indeed, the reward function can be changed at any time during or in-between episodes, without any need for additional training. While vanilla CEM suffers from poor data-efficiency in high-dimensional spaces, recent efforts \cite{pinneri2021sample} demonstrate that this aspect can be significantly improved, in particular via sampling time-correlated action sequences. We use CEM-based planning in some of our experiments, and in others, further simplify action selection by sampling from a discrete set of velocity controls. 

  \subsection{Irregular/asynchronous actions and observations.}
  \label{sec_related_irreg}

  The vast majority of the RL literature considers discretized time-steps with regular observations and actions, and most prior work on continuous time systems make simplifying assumptions such as partially known and/or linearized dynamics systems \cite{lee2021policy, abu2005nearly, wang2020reinforcement}, or are applied in model-free settings \cite{tallec2019making,ghavamzadeh2001continuous}. However, more recent efforts build on Neural Ordinary Differential Equations (N-ODEs) which outperform discrete approaches to modeling continuous-time dynamics \cite{chen2018neural}. In their work, Du et al. \cite{du2020model} model environment dynamics using N-ODEs in order to learn policies in semi-MDP problem settings, while the continuous-time RL framework of Yildiz et al. \cite{yildiz2021continuous} is notable for providing state uncertainty estimates. Although our use of N-ODEs is similar to what was done in those works, it should be noted that our focus is on an orthogonal direction: our aim is to investigate the combination of meta-learning and N-ODEs in order to simultaneously address the problem of irregular/asynchronous actions and observations and discontinuous changes in system dynamics on a per episode basis. This distinction is reflected in our experimental setup, which is closer to real-world applications. \color{black} We also note that several recent works from the control literature have also incorporated neural ODEs \cite{chee2022knode, jiahao2023online,richards2022control} which equips them for dealing with irregular inputs\footnote{\color{black} We note in passing that this ability is more often than not a by-product of their methodology, as their focus is essentially on accurate dynamics modeling rather than on integrating data from different sources that is produced at different rates (\textit{e.g.} from asynchronous publication to various \texttt{ros} topics), which is our principal motivation.\color{black}}. The first two of these methods aim at increasing accuracy in a single environment, and while the two others \cite{jiahao2023online, richards2022control} do target adaptivity to different environments, they both \textemdash as we will discuss in more detail in section \S\ref{sec_related_system} \textemdash  are limited to specific types of tasks and trade-off flexibility for accuracy.\color{black}

  \subsection{Meta-learning.} 
  \label{sec_related_meta}
  Leveraging previous experience to improve the learning process has been extensively studied in the meta-learning \cite{hospedales2021meta, Schaul:2010} literature. As even when considering the most restrictive definitions of meta-learning in which identical train/test conditions, end-to-end optimization and sample splitting are essential \cite{hospedales2021meta}, one is left with a plethora of methods that differ, among other things, in what meta-parameters they optimize. For example, many meta algorithms learn a recurrent network that models policies \cite{duan2016rl} or optimizers/update rules \cite{oh2020discovering}, while others optimize hyperparameters \cite{xu2018meta} or metrics \cite{vinyals2016matching}. It is thus important to note that the meta-parameter that we seek to optimize is a prior dynamics model (more precisely, a neural ODE), which when used as an initialization in novel control tasks, should be able to quickly adapt to represent the dynamics of that environment. Learning adaptive priors is precisely what algorithms in the MAML family \cite{finn2017model, song2019maml} are designed for. In particular, we update our parameters in a manner similar to ES-MAML \cite{song2019maml}, as it alleviates the need for the estimation of second order.

  \subsection{Adaptive MB-RL / MPC}
  \label{sec_related_system}
\color{black}
  Controlling an agent in a manner that is adaptive to different tasks and/or environment dynamics has been the subject of several works in the RL literature. For example, the work of Sekar \textit{et al.} builds a global world model that is used in order to adapt to unseen downstream tasks. Closer to our work is the method proposed by Nagabandi \textit{et al.} \cite{nagabandi2018learning}, which uses MAML-based meta-learning to adapt the model to novel environment dynamics in a manner that is not dependent on a particular reward function. Other works have extended PETS\cite{chua2018deep} with a meta-learning component \cite{belkhale2021model}. More recently, transformer-based in context-meta learning in the style of RL$^2$ \cite{duan2016rl} has been coupled with model-based RL in order to train an RL agent at scale (akin to an early foundation model for RL). While many of those approaches do indeed result in remarkable adaptivity, they all assume the classical RL settings in which actions and observations are synchronized.

  On the other hand, some recent MPC methods that aim to learn an adaptive control process \cite{jiahao2023online, richards2022control} leverage neural ODEs to model the underlying dynamic systems. While they are therefore naturally equipped to handle irregular inputs, they are less flexible than their RL-based counterparts. In particular, the model learned in the work of Jiahao \textit{et al.}\cite{jiahao2023online} results from adapting a weighted average over a history of learned models, which results in gradual forgetting of previous environments. While the method proposed by Richards \textit{et al.}\cite{richards2022control} results in a more versatile controller, it is supervised by a trajectory-tracking meta-objective, which reduces its flexibility in terms of adaptation to other tasks, such as exploration. Furthermore, online adaptation in their work consists in forward simulation using meta-parameters and approximate dynamics that are all learned off-line. While this is an efficient approach, its adaptivity is limited by the trajectory-environment space that is covered by the offline dataset.
\color{Black}

\section{ACUMEN}
  \label{sec_acumen}

  We propose an algorithm based on Model Predictive Control (MPC) \cite{pinneri2021sample, moerland2020model} with a learned and dynamically adapted environment model based on the formalism of Neural Ordinary Differential Equations (N-ODE) \cite{chen2018neural}. In order to ensure performance on unseen tasks, learning experience from previous episodes is used to maintain a prior on the weights of the N-ODE with which each new episode is initialized. Each of the following subsections is dedicated to one of those components. \color{black} A high-level overview of our approach is given in figure \ref{fig_high_level}, and pseudo-code is given in algorithm \ref{algo_pseudo_main} \color{black}.

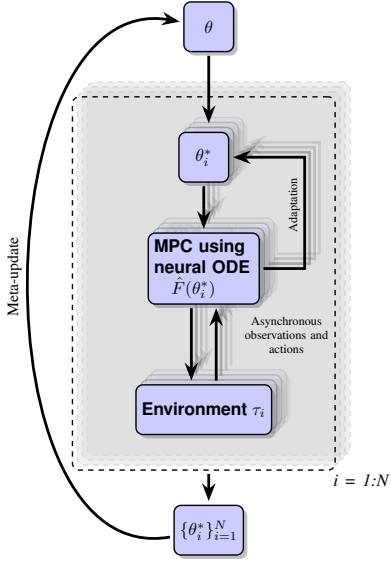
\begin{figure}[ht]
  \centering
 \resizebox{0.65\textwidth}{!}{
\begin{tikzpicture}[
  box/.style={
    draw,
    minimum size=1cm,
    rounded corners,
    thick,
    fill=blue!20,
    font=\sffamily\bfseries,
    drop shadow
  },
  arrow/.style={
    ->,
    ultra thick,
    >=Stealth,
    shorten >=2pt,
    shorten <=2pt
  },
  module/.style={
    draw,
    dashed,
    rounded corners,
    thick,
    inner xsep=36pt,
    inner ysep=20pt,
    fit={(A) (C)},
    label={[anchor=north]above:}
  }
]
  \foreach \i/\j in {3mm/3mm, 2mm/2mm, 1mm/1mm} {
    \begin{scope}[xshift=\i, yshift=\j, opacity=0.4]
      \node[box] (A) at (0, 0) {$\theta_i^*$};
      \node[box, below=1cm of A,text width=2.0cm] (B) {MPC using\\ neural ODE \\ \ \ \ $\hat{F}(\theta_i^*)$};
      \node[box, below=1.6cm of B] (C) {Environment $\tau_i$};
      \draw[arrow] (A) -- (B);
      \draw[arrow] ([xshift=-0.25cm]B.south) -- ([xshift=-0.25cm]C.north);
      \draw[arrow] ([xshift=0.25cm]C.north) -- ([xshift=0.25cm]B.south);
      \draw[arrow] (B.east) -- ++(0.9cm, 0) |- (A.east);
      \node[module, draw=gray!70, fill=gray!30] {};
    \end{scope}
  }

 \node[box, above=1.5cm of A] (I) {$\theta$};
  \node[box, below=1.5cm of C] (O) {$\{\theta_i^*\}_{i=1}^N$};
  \draw[arrow] (O) to[out=190, in=-190] node[midway, left,align=center,font=\small, xshift=-0.3cm, yshift=1.0cm, rotate=90] {Meta-update} (I);

  \node[fit={(A) (C)}, inner sep=20pt] (S) {};

  \path[name path=arrowPath] (O) -- (C);
  \path[name path=moduleBoundary] (S.north west) -- (S.north east) -- (S.south east) -- (S.south west) -- cycle;
  \path[name intersections={of=arrowPath and moduleBoundary, by=intersectionPoint}];
  \draw[arrow] ([yshift=-0.1cm]intersectionPoint) -- (O);

  \draw[arrow] (I) -- (A);


  \node[box] (A) at (0, 0) {$\theta_i^*$};
  \node[box, below=1cm of A,text width=2.0cm] (B) {MPC using\\ neural ODE \\ \ \ \ $\hat{F}(\theta_i^*)$};
  \node[box, below=1.6cm of B] (C) {Environment $\tau_i$};
  \draw[arrow] (A) -- (B);
  \draw[arrow] ([xshift=-0.25cm]B.south) -- ([xshift=-0.25cm]C.north);
  \draw[arrow] ([xshift=0.25cm]C.north) --  node[midway, left,align=center,font=\scriptsize, xshift=2.4cm, yshift=0.2cm] {Asynchronous\\observations and\\actions} ([xshift=0.25cm]B.south);
  \draw[arrow] (B.east) -- ++(0.9cm, 0)  node[midway, left,align=center,font=\scriptsize,rotate=90, xshift=1.98cm, yshift=-0.24cm] {Adaptation} |- (A.east);
  \node[module] {};

  \node[below right=0.05cm and 0.3cm of S.south east, font=\itshape] {i = 1:N};
\end{tikzpicture}
}

\caption{\small{A high-level view of the proposed algorithm. At each meta-iteration, $N$ different environments $\tau_1, ..., \tau_N$ are sampled and each $\tau_i$ is paired with a neural ODE (noted $\hat{F}(\theta_i^*)$ in the figure). Episode $i$ then consists in a sampling-based model-predictive pipeline in environment $\tau_i$ (see sections \ref{subsec_mpc} and algorithm \ref{algo_action_sel}). During such an episode, $\hat{F}(\theta_i^*)$ is used as and approximate world model for state predictions, and $\theta_i^*$ is continuously refined in order to improve the predictions of $\hat{F}(\theta_i^*)$. After all $N$ episodes have been completed (either in parallel or sequentially), the optimized $\{\theta_i^*\}_{i=1}^N$ are used to estimate the gradient for the meta-update (equations \ref{eq_meta}, \ref{eq_meta_2} in the main text, and line $30$ in algorithm \ref{algo_pseudo_main}).}}
   \label{fig_high_level}
\end{figure}

\begin{algorithm}
  \SetCommentSty{mycommfontblue}
  \begin{footnotesize}
    \KwIn{Environment distribution $\mathcal{P}(\tau)$, number of environments to sample for each meta update $N$, train/validation split ratio $r_{split}\in (0,1)$, maximum number of iterations for inner loop (neural ODE) optimizations $N_{it}$, optionally prior parameters of the N-ODE $\theta_0$, hyperparameters \texttt{action\_selection\_hyperparams} (see algorithm \ref{algo_action_sel}), function $\xi$ mapping sliding windows of observations to state approximations, size of sliding window of observations $M$, learning rate for the meta update $\alpha$, standard deviation $\sigma$ for the estimation of the meta update}
    \KwOut{prior parameters to use for future sessions $\theta$}
    \ \\
    $\theta\leftarrow \texttt{RandomInit()}$ if $\theta_0$ \textrm{is not given;} else $\theta \leftarrow \theta_0$ \\
    \SetAlgoLined
    \While {\texttt{\textrm{session no over}}}
    {
      \tcp{sample i.i.d vectors for the ES update of the meta-parameters}
      $\epsilon_1, ..., \epsilon_N \sim \mathcal{N}(\textbf{0},\textbf{I})$\\
      \For {$i \in \{1, ..., N\}$ }
      {
        $\tau_i \sim \mathcal{P}(\tau)$\\
        $\mathcal{D}_i^{train}\leftarrow \emptyset$\\
        $\mathcal{D}_i^{val}\leftarrow \emptyset$\\
        $\theta_i^*\leftarrow \theta + \epsilon_i\sigma$\\
        $C_{hist}\leftarrow \emptyset$\tcp{history of applied actions}
        \While {\textrm{not} ($\tau_i$.\texttt{success} or $\tau_i$.\texttt{timeout})}
        {
          \ \\ 
          \tcp{get the $M$ most recent observations}
          $\omega_1, ..., \omega_M \leftarrow \tau_i$.\texttt{GetSlidingWindow({$M$})}\\ 
          \tcp{get all actions sent since the oldest observation in the window}
          $c_1, ..., c_l \leftarrow$\texttt{GetAllActionsSince(}$\omega_1.$\texttt{timestamp}, $C_{hist}$)\\ 
          \ \\
          \tcp{Map the sliding window to a state approximation}
          $\hat{z} \leftarrow \xi(\omega_1,...,\omega_M)$\\
          
          \ \\
          \tcp{Choose the next action sequence (see algorithm \ref{algo_action_sel} for details)}
          \texttt{seq}$\leftarrow$ \textrm{\texttt{SelectNewCommand(}}$\theta_i^*$, $\hat{z}$,$\{c_1,...,c_l\}$,\texttt{action\_selection\_hyperparams})\\
          
          \tcp{Apply the $K$ first actions in the sequence}
          $\tau_i.$\texttt{Apply(seq}$_{1:K}$\texttt{)}\\
          
          \tcp{Add the applied actions to command history}
          $C_{hist}.$\texttt{Append(seq}$_{1:K}$\texttt{)}\\
          
          \ \\ 
          \tcp{split gathered data into train/validation sets}
          $d_{train}, d_{val} \leftarrow$ \texttt{SampleSplit}$([\omega_1, ..., \omega_M], [c_1, ..., c_l],r_{split})$\\
          $\mathcal{D}_i^{train}.$\texttt{Append}($d_{train})$\\
          $\mathcal{D}_i^{val}.$\texttt{Append}($d_{val})$\\

          \tcp{optimize the prior if necessary (see appendix \ref{appendix_loss} for details on OptimizeNODE)}
          \If {\texttt\textrm{train\_freq}}
          {
            $\theta_i^*\leftarrow$\texttt{OptimizeNODE}$(\theta_i^*, \mathcal{D}_i^{train},N_{it})$\\
          }
        }
      }
      \tcp{update the meta-parameters $\theta$}
      $\theta\leftarrow \theta - \frac{\alpha}{N\sigma}\sum_{i=1}^{N} \mathcal{L}_i({\theta}_i^*, \mathcal{D}_i^{val})\epsilon_i$
    }
    \caption{\small{The ACUMEN algorithm.}}
    \label{algo_pseudo_main}
  \end{footnotesize}
\end{algorithm}

 \begin{algorithm}
  \SetCommentSty{mycommfontblue}
  \begin{footnotesize}
    \KwIn{Neural ODE weights $\theta$, state approximation $\hat{z}(t_0)$ at time $t_0$, desired duration of state propagation $\Delta t$, previous actions $c_1,...,c_l$ (augmented with their timestamp info), number of action sequences $N_p$ to sample, number of elite action sequences $N_e$, length of action sequence to sample $H$, reward function $R(.)$, prior distribution on actions $P_{\psi}(A)$ (parametrized by $\psi$), initial parameters $\psi_0$ for $P_{\psi}(A)$, convergence criterion $\mathcal{Y}$}
    \KwOut{action sequence $a^*_1,...a^*_H$} 
    \SetAlgoLined
    \DontPrintSemicolon
    \SetKwFunction{FMain}{SelectNewCommand}
    \SetKwProg{Fn}{Function}{:}{}
    \Fn{\FMain{$\theta$, $\hat{z}(t_0)$, $\Delta t$, $\{c_1,...,c_l\}$, $N_p$, $N_e$, $H$, $R(.)$, $\mathcal{Y}$}}{
      $\psi\leftarrow\psi_0$\\
      \tcp{initialize elite set}
      $\texttt{elites}\leftarrow\emptyset$\\
      \While {\texttt{not $\mathcal{Y}$}}
      {
        \tcp{sample $N_p$ action sequences of length $H$ at regular intervals in $[t_0, t_0+\Delta_t]$}
        $a^1_{1:H}, ..., a^{N_p}_{1:H}\sim P_{\psi}(A)$\\
        \For {\texttt{each} $a^{i}_{1:H}$}
        {
          $\pi_i\leftarrow[c_1,...,c_l]$\texttt{.concatenate(}$a^i_{1:H})$)\\
          \tcp{Let $u_i(t)$ the function that computes actions via interpolating elements of $\pi_i$}
          $u_i\leftarrow\mathcal{I} \circ \pi_i$\\
          \tcp{propagate the state}
          $\hat{z}_i\leftarrow\hat{z}(t_0)+\int_{t_0}^{\Delta t} \hat{\mathcal{F}}(\hat{z}(t),u_i(t),\theta)dt.$\\ 
          \tcp{compute associated reward}
          $r_i\leftarrow$R($\hat{z_i}$)
        }
        $\texttt{elites}\leftarrow$\texttt{select best }$N_e$\texttt{ action sequences according to the }$r_i$\\
        \tcp{Update prior distribution $P_{\psi}(A)$ on commands}
        $\psi \leftarrow$\texttt{updatePriorDistrib}($a^1_{1:H},...,a^{N_p}_{1:H}, r_1, ..., r_{N_p}$)\\
      }
      \tcp{return the best action sequence (alternatively, re-sample from $P_{\psi}(A)$)}
      \KwRet $\texttt{elites}$
    }
    \caption{\small{An action selection step using the learned neural ODE. While predicted actions $a^i_{1:H}$ are regularly sampled, their application on the system will be irregular. Thus, the history of actions $c_1,...,c_n$ will be composed of irregularly applied actions, hence the advantage of using the neural-ODE formalism. Note that while the action sampling process outlined here includes CEM-based sampling with time-correlated actions (as in \cite{pinneri2021sample}) as a special case, it can also be reduced to a simple random shooting depending on the choice of $P_{\psi}(A)$ and other hyperparameters. We will adjust those on a per application basis in our experiments. See sections \ref{sec_experim_simu}, \ref{sec_experim_soto} and appendix \ref{appendix_cmd_sel} for more details.}}
    \label{algo_action_sel}
  \end{footnotesize}
\end{algorithm}

  \subsection{Handling irregular/asynchronous observations and actions}
  \label{subsec_nodes}
  As the state in the problem settings that we discussed in section \S\ref{formulation_sec} is partially observable, we assume the existence of a learned or hand-designed function $\mathcal{\xi}_n: \omega_0 \times \omega_1 \times ... \times \omega_n \rightarrow Z$ that maps a sliding window of observations $W=\{\omega\}_{i=1}^n$ to a state approximation, \textit{i.e.} $\xi(W)=\hat{z} \in Z$. When clear from context, we will drop the window size from the notation and simply write $\xi$. We then use a neural ODE $\hat{\mathcal{F}}$, parametrized by weights $\theta$ to approximate the dynamics evolution function $\mathcal{F}$:

 \begin{equation}
   \hat{z}(t_0+\Delta t)\approx\hat{z}(t_0)+\int_{t_0}^{\Delta t} \hat{\mathcal{F}}(\hat{z}(t),t,u(t),\theta)dt.
  \end{equation}
  \label{eq_def}

  As we are interested in irregular actions, the function $u(.)$ will in effect be the composition of two functions: a decision process $\pi(t_0-s, t_0+\Delta t)$ that outputs irregular actions at discrete timestamp falling in $[t_0-s, t_0+\Delta t]$, and an interpolation function that interpolates those actions to produce action values at continuous timestamps\footnote{\color{black}In our work, $\pi$ is not a trained policy, but is instead sampled from a distribution $P_{\psi}$. To keep notations concise, we have omitted the dependency of $\pi$ on that distribution in the notation $\pi(t_0-s, t_0+\Delta t)$.\color{black}}. More formally, $u$ can be written as

  \begin{equation}
    u(t)= (\mathcal{I} \circ \pi(t_0-s,t_0+\Delta t))(t) \ \ \ \ \forall t \in[t_0-s, t_0+\Delta t] 
  \end{equation}

  for some (small) real value $s$. In our work, the interpolation $\mathcal{I}$ is the linear interpolation function.

  In our action-sampling based system (algorithm \ref{algo_action_sel}), $N_p$ different action sequences, each of length $H$, are sampled and considered as possible continuations of the current history of actions. Each of these action sequences can be thought of as a decision process $\pi_i$ (with $i \in \{1,...,N_p\}$). Since the interpolation function $\mathcal{I}$ is fixed, it is equivalent to say that we samples $N_p$ functions $u_{i}\triangleq \mathcal{I}\circ \pi_i$.

\begin{figure}[ht!]
  \centering
  \captionsetup[subfigure]{justification=centering}
    \subfloat[][]{
      \includegraphics[width=70mm,trim={0cm 0 0cm 0.0cm},clip]{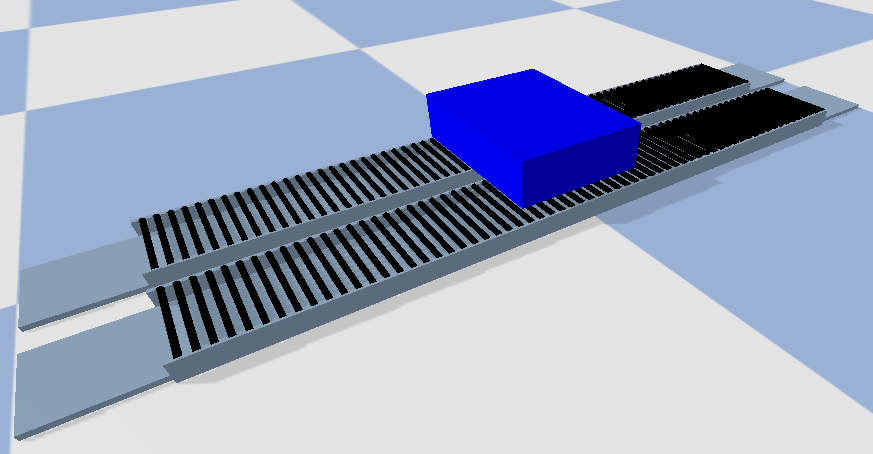}}\\
    \subfloat[][]{
      \includegraphics[width=22mm,trim={0cm 0 0cm 0.0cm},clip]{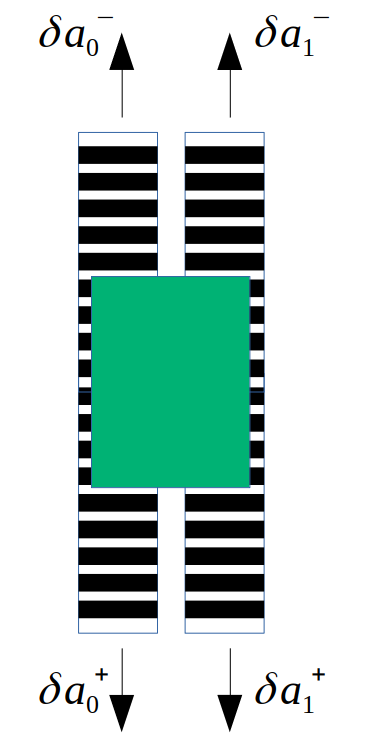}}
    \hspace*{1.0cm}
 \subfloat[][]{\raisebox{2ex}{
     \includegraphics[width=32mm,trim={0cm 0 0cm 0.0cm},clip]{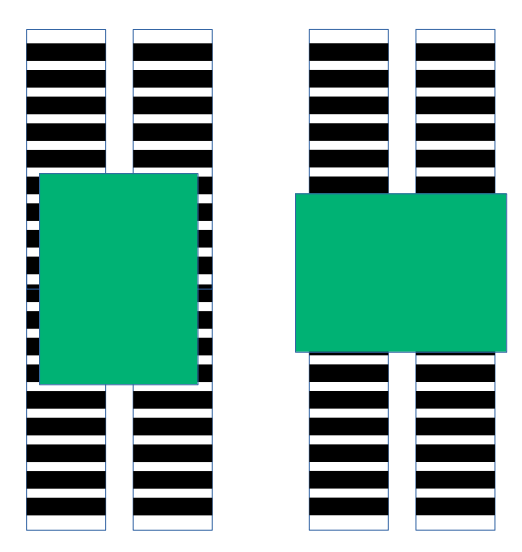}}}
   \caption{\small{\textbf{(a)} Screenshot of the developed robot simulation, based on simulating two roller conveyors. \textbf{(b)} Schematic view of the system, showing a green box on top of two conveyor belts. The conveyor belts are independent, and the motion of each one is controlled via a one-dimensional velocity command in $[-1,1]$. Positive and negative actions, \textit{e.g.} $\delta a^+_0,\delta a^-_0$ respectively move the left conveyor in the top or bottom direction. Commands such as $\delta a^+_1,\delta a^-_1$ should affect the right-hand conveyor belt in similar fashion. \textbf{(c, left)} Illustration of the initial conditions in each episode. \textbf{(c, right)} Target pose that we aim to reach through controlling the conveyor belts. Note that the mass and inertial properties significantly and discontinuously vary in-between episodes.}}
   \label{fig_simulation_description}
\end{figure}

  \subsection{Planning with a continuously updated model}
  \label{subsec_mpc}

  The proposed method is based on Model Predictive Control (MPC) with a learned model, that is updated throughout the duration of an episode. At each iteration of the algorithm (lines 11-28 in algorithm \ref{algo_pseudo_main}), a sliding window consisting of the last $M$ observations $\{\omega_1,...,\omega_M\}$ is mapped to a state estimation $\hat{z}(t)$. Actions $\{c_1,...,c_l\}$ (in general, $l\neq M$) that have been applied since the time at which $\omega_1$ was observed are gathered. All observations and actions are added to a buffer with the aim of updating the model.

We determine an action sequence $a^*_{1:H}$, that when appended to $\{c_1,...,c_l\}$, would result in the most promising future state according to the predictions of the learned model. The sampling and selection process is detailed in algorithm \ref{algo_action_sel}. As previously discussed in detail in section \S\ref{sec_related}, our decision to rely on command sampling is not only motivated by its demonstrated efficacy and robustness, but also by its simplicity. 

Once a sequence $a^*_{1:H}$ has been selected, its first action is applied on the system. The learned model is updated every \textit{\texttt\textrm{train\_freq}} iterations. Note that the loss function for training the neural ODE and the dense reward function $R(.)$ are both application dependent. See appendices \ref{appendix_loss} and \ref{appendix_cmd_sel} for details about those used in our experiments.

  \subsection{Learning a neural ODE prior}
  \label{subsec_metalearning}

Our objective of obtaining a point-wise estimate of weights that could serve as an adaptive prior in environments with previously unseen dynamics can be formulated via the following equation:

\begin{equation}
  \theta\leftarrow\theta-\alpha \nabla\sum_{i=1}^N\mathcal{L}(\mathcal{T}^i_{val}, \theta-\lambda \nabla\mathcal{L}(\mathcal{T}^i_{train},\theta))
  \label{eq_meta}
  \end{equation}

\begin{figure*}[ht!]
  \centering
  \captionsetup[subfigure]{justification=centering}
    \subfloat[][]{
      \includegraphics[width=46mm,trim={0cm 0 0cm 0.0cm},clip]{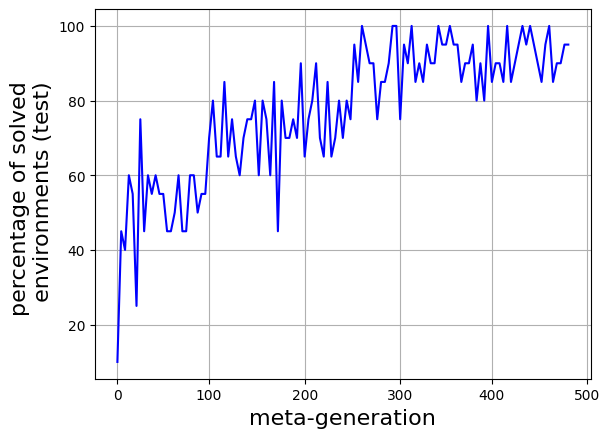}}
    \subfloat[][]{
      \includegraphics[width=46mm,trim={0cm 0 0cm 0.0cm},clip]{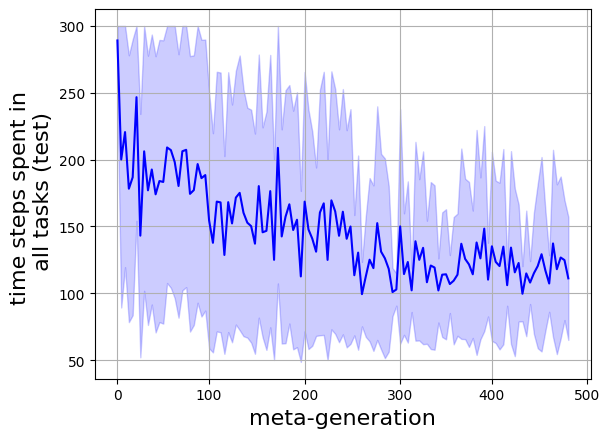}}
 \subfloat[][]{
      \includegraphics[width=46mm,trim={0cm 0 0cm 0.0cm},clip]{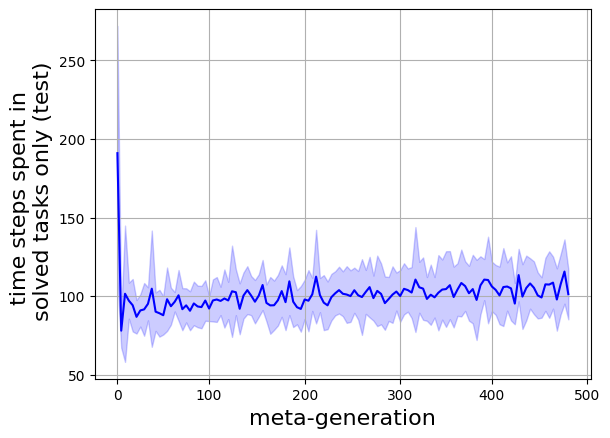}}\\
    \caption{\small{\textbf{(a)} The evolution of the percentage of solved environments among the $M_{test}=20$ ones that are sampled from the test distribution $P(\mathcal{T}_{box})$ at each meta-generation, starting from a random neural ODE. \textbf{(b)} In this figure, the value displayed at each meta-generation is the time spent in all environments (solved and unsolved) that have been sampled at this meta-iteration. \textbf{(c)} Same as in \textbf{(b)}, but for \textit{solved} environments only. Notice the large drop in the number of necessary timesteps in the $\sim 10$ first meta-generations.}}
   \label{fig_complete_system}
\end{figure*}

\begin{figure*}[h!]
  \centering
  \captionsetup[subfigure]{justification=centering}
    \subfloat[][]{
      \includegraphics[width=46mm,trim={0cm 0 0cm 0.0cm},clip]{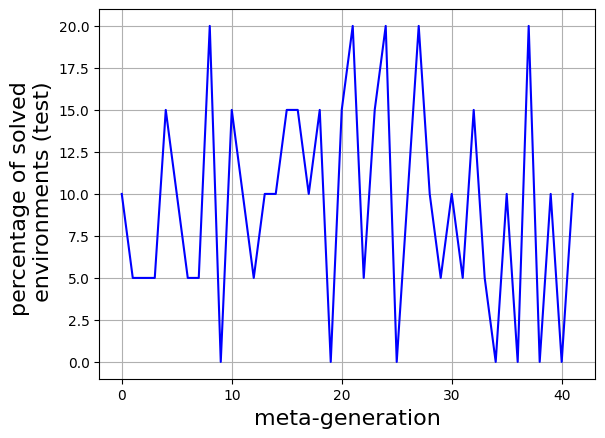}}
    \subfloat[][]{
      \includegraphics[width=46mm,trim={0cm 0 0cm 0.0cm},clip]{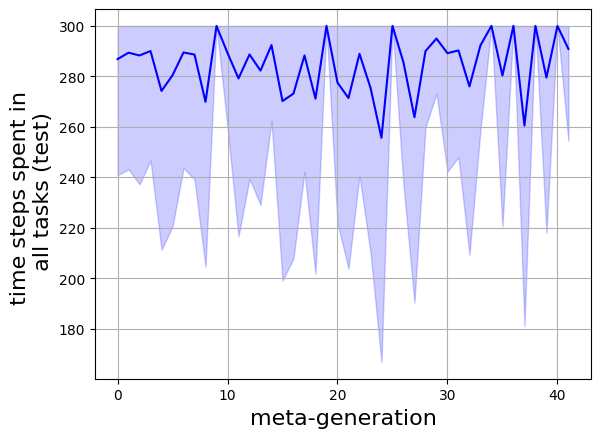}}
 \subfloat[][]{
      \includegraphics[width=46mm,trim={0cm 0 0cm 0.0cm},clip]{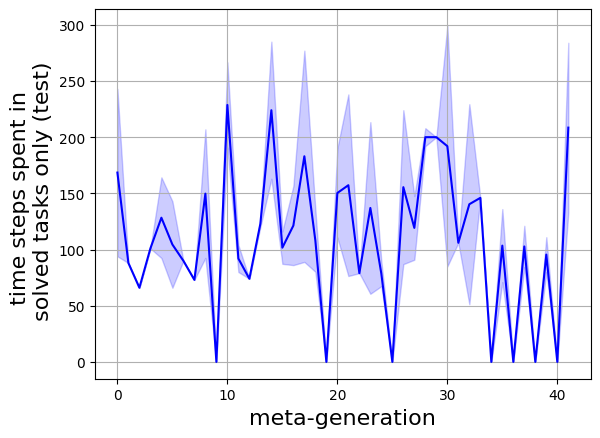}}
    \caption{\small{Result of meta-update ablation. \textbf{(a)}. It can be seen that a random neural ODE can be adapted to solve an average of $\sim 9\%$ of the sampled environments in the given time limit of $H_{max}=300$. \textbf{(b)} Naturally, without the meta-update, there is no reduction in the average time taken to solve the environments. \textbf{(c)} The average and the variance of time-steps spent across \textit{solved} environments only. Note that \textit{values of zero indicate that none of the environments were solved}. The remaining values are in general larger than those that were reached at convergence with the meta-update enabled (figure \ref{fig_complete_system}(c)).}}

   \label{fig_meta_update_ablation} 
\end{figure*}

  in which $\mathcal{L}(.,\theta)$ denotes the training loss of a neural ODE on some dataset. As is classically done in the meta-learning literature for the sake of conciseness in notations, each of the inner and outer optimization problems have been written as a single gradient descent update. However, in practice the number of updates is arbitrary.

Notice that the inner optimizations, \textit{i.e.} $\theta-\lambda \nabla\mathcal{L}(\mathcal{T}^i_{train},\theta)$, correspond to model-predictive episodes (lines 11-28 in algorithm \ref{algo_pseudo_main}).

Computing the gradient update for the outer level optimization problem requires differentiating through the gradient computed in the inner optimization. As this would necessitate computing higher order derivatives of neural ODEs, we simplify the outer level update by using an estimator based on Evolution Strategies \cite{salimans2017evolution}:

\begin{equation}
  \begin{split}
    \theta &\leftarrow \theta - \frac{1}{\sigma}\mathbb{E}_{\tau\sim P(\tau),\epsilon \sim \mathcal{N}(0,I)}[ \\
    &\mathcal{L}(\mathcal{T}^{\tau}_{val}, \theta+\epsilon-\lambda \nabla\mathcal{L}(\mathcal{T}^{\tau}_{train},\theta+\epsilon))\epsilon].
  \end{split}
  \label{eq_meta_2}
\end{equation}

Note that this update, approximated using $N$ sampled environments in algorithm \ref{algo_pseudo_main} (line 30) is equivalent to the zero-order ES-MAML update \cite{song2019maml}.

\section{Experimental validation in simulated environments}
\label{sec_experim_simu}

We first present experiments on two simulated robots. The first one is inspired by the SOTO2 robot manufactured by Magazino Gmbh, and uses two conveyor belts to manipulate boxes of different inertial properties. As the simulation is developed in pybullet \cite{coumans2021}, command and observation irregularities are simulated by randomly dropping observations and by applying interpolated commands at random timestamps. The second experiment is based on the Gazebo Turtlebot3 simulation\cite{gazeboturtle}. As communicating with the Gazebo simulator is done via ROS topics and services, the observations and actions already suffer from some irregularity that we further accentuate by dropping random observations in order to better highlight the advantages of using neural ODEs. For both robots, the physical properties of the environment are randomly sampled at the beginning of an episode.

In addition to evaluating the ability of the ACUMEN algorithm to jointly address asynchronous/irregular actions and observations as well as discontinuous changes in environment dynamics, we also highlight the effect of its different components through ablation studies. In particular, we highlight the advantages of using neural-ODEs. Classical model-based RL approaches usually learn a model of the form $\Delta z_{t+1}=M(z_t,a_t)$, which corresponds to (PO)MDPs. A simple extension to that model for the case of irregular/asynchronous actions can be obtained by considering a recurrent neural network which would take as input the sequence of tuples $(a_0, \Delta t_0), ... ,(a_n, \Delta t_n)$ with $\Delta t_i$ the elapsed time between $a_i$ and $a_{i+1}$ for $i\neq n$, and $\Delta t_n$ the elapsed time between $a_n$ and the next observation (or during planning, the next predicted state). In this paper, we implement this model as a vanilla stacked RNN. Note that this is a slight generalization of state propagation of the form $z_t=z_{t-1}+M(z_{t-1},a_{t-1},\Delta t_{t-1})$, \color{black} and that the latter \color{black} has been shown in prior work \cite{yildiz2021continuous} to be inferior to neural ODEs in continuous time settings. In the rest of the paper, we will use the terms RNN and stacked RNN and recurrent model interchangeably when the context allows it.

Details regarding loss and reward functions as well as various hyper-parameters can be found in the appendixes.

\subsection{Simulated box rotation using conveyor belts}
\label{sec_experim_simbox}

This robotic simulation is inspired by the SOTO2 robot manufactured by Magazino Gmbh. Before presenting our results, we detail the simulation as well as some algorithmic choices in the following section. Results from the complete system are presented in section \ref{simulation_system_results}, and ablation studies follow in sections \ref{sec_ablation}.

\subsubsection{Simulated Box rotation: experiment description}
\label{sec_sim_description}

The simulation, developed using the bullet physics engine \cite{coumans2021}, defines the following problem: given two parallel and independently controlled conveyor belts separated by some distance (figure \ref{fig_simulation_description}(a)), the aim is to rotate a parcel by $\frac{\pi}{2}$ radians. The default problem setting is illustrated in figure \ref{fig_simulation_description}(c). At the beginning of an episode, a box of varying inertial properties is sampled from a distribution $P(\mathcal{T}_{box})$, and is positioned (plus or minus some gaussian noise) at the pose shown in figure \ref{fig_simulation_description}(c, left). The task is considered solved if the system reached (with some tolerance) the state given in the right hand of figure \ref{fig_simulation_description}(c). The environment, shown in \ref{fig_simulation_description}(b), is based on simulated roller conveyors \cite{mcguire2009conveyors}.

The box distribution $P(\mathcal{T}_{box})$ is defined by a random choice of mass in the interval $[0.1kg, 3kg]$ and a mass distribution given by random $3d$ Gaussian distributions, which can have dense covariances. To simplify matters, we use ground truth $6d$ box poses as observations. Given the last two observed box poses $\omega_{t-1}, \omega_t$, we approximate the state to propagate as $\hat{z}=\xi(\omega_t, \omega_{t-1})\triangleq[\omega_t, \frac{\omega_t-\omega_{t-1}}{\Delta t}]$. Note that the action space is two-dimensional and defined by \begin{scriptsize}$[-1,1]\times [-1,1]$\end{scriptsize}, as each of the conveyors can receive velocity controls independently of the other one. As specified in appendix \ref{appendix_cmd_sel_pybullet}, the distance from a given pose and the target pose is used as a dense reward signal to guide the action selection process. The action sampling mechanism that is used is a particular instantiation of algorithm \ref{algo_action_sel} which is equivalent to the CEM-based control method with colored noise from \cite{pinneri2021sample}. Details are given in appendix \ref{appendix_cmd_sel}.

For benchmarking purposes, we define each timestep as the time elapsed between applying actions in the physics engine, which is done at regular intervals. We however emphasize that those actions are not used by our control algorithm, which only receives actions that the result of linear interpolations at random time-steps. The maximum number of timesteps for each episode is set to $H_{max}=300$.

In all experiments, the \texttt{RK45} solver \cite{dormand1980family} (commonly referred to as \texttt{dopri5}) is used. 

\noindent \textbf{Irregular/asynchronous observations and actions.} We simulate irregular/asynchronous observations and actions by 1- Returning only one random observation among $K_s$ successive observations, 2- returning actions at randomly interpolated time-stamps in between real actions and 3- Discarding selected observations with some probability $\eta$. In the following experiments, a value of $K_s=3$ has been chosen, and $\eta=0.05$. The colored noise parameter used in CEM-based sampling (algorithm \ref{algo_pseudo_cem}) was set to $\beta=2$.

\subsubsection{Simulated box rotation: system results}
\label{simulation_system_results}

The objective of this subsection is to validate the complete system in the presented simulation, where irregular/asynchronous actions and observations have been simulated, and where environment dynamics vary from one episode to the next due to the random sampling of box mass and inertial properties.

A neural ODE is initialized with random weights, and receives successive meta updates in order to form the learned prior. In each meta iteration, $N=20$ environments $\tau_1, ..., \tau_N$ are sampled from the environment distribution. As detailed in algorithm \ref{algo_pseudo_main}, the neural ODE is independently adapted to each of the $\tau_i$ using data gathered during the episode, the validation split of which is then used in the meta update. 

The result of the experiment are plotted in figure \ref{fig_complete_system}. As figure \ref{fig_complete_system}(a) shows, at initialization, model-predictive control with the random neural ODE is only able to solve about $5\%$ of the environments that are sampled from the test distribution. In this first meta iteration, the variance of episode lengths (figure \ref{fig_complete_system}(b)) is large, with most episode requiring many updates to the neural ODE. Hence, the large mean episode length reported in that same figure. 

It can be seen that as the learned prior is optimized during successive meta-iteration, the percentage of successfully solved test environments increases until eventually reaching a point where it oscillates between $95-100\%$ (figure \ref{fig_complete_system}(a)). Simultaneously, the number of steps spent across all environments (solved and unsolved) decreases as the optimization of the prior progresses (figure \ref{fig_complete_system}(b)). 

Figure \ref{fig_complete_system}(c) shows the number of timesteps spent in environments that the system was able to solve at a given meta iteration. After an initial large drop in mean and variance in the first few ($<10$) meta iterations, the number of necessary steps for solving an environment stabilizes to around $100$ timesteps, while the number of success (figure \ref{fig_complete_system}(a)) continues to grow. A possible explanation for this phenomenon comes from observing the behavior of models specialized for an environment, an example of which will be discussed in the next subsection. Models that are specialized to an environment result in near-optimal state-action trajectories connecting the initial box pose to the target pose, and take less than $50$ timesteps \footnote{The number of timesteps necessary for an episode based on a specialist network to succeed was computed by sampling $N_{env}=30$ environments and training $N_{env}$ corresponding neural ODEs on each of them to obtain specialist networks. Then, each neural ODE was used in $10$ control episodes on the environment it was trained on. Except for a few outliers, the average number of timesteps fell in between $40$ and $50$.}. A general prior would then be positioned in some area of parameter space minimizing its average distance to several specialist networks, hence the higher number of updates necessary for its adaptation. In addition to that, we note that our meta-update does not impose hard constraints on the adaptation speed.

\subsubsection{Simulated box rotation: Ablations}
\label{sec_ablation}
Our principal aim in this paper is to investigate the advantages of combining meta-learning and neural-ODEs in the contexts that were previously discussed. Therefore, we consider two ablation experiments. In the first one, we consider two different manners in which the meta-update component can be removed. In the second ablation study, we compare a specialized neural ODE to a specialized Recurrent network.

\textbf{Meta-update ablation}. We first naively disable the meta-update in ACUMEN. The results of this experiment are reported in figure \ref{fig_meta_update_ablation}. Unsurprisingly, without the meta update, the adaptations that the neural ODE goes through within distinct episodes is only sufficient for solving on average about $\sim9\%$ of the sampled environments in the given time limit of $H_{max}=300$ (which is the same as in all our simulation-based experiments). The number of timesteps solved in all environments (solved or unsolved) displays significantly higher mean and variance.

To further investigate the advantages of the learned prior, we compare its performance with that of a specialist model trained until convergence on a single "average" environment, henceforth noted $E_1$, given by a box with mass $1.5kg$ and uniform mass distribution. The results are compiled in table \ref{table_specialise}. As expected, the learned prior is able to adapt to and solve $91.3\%$ of the environments within the (per episode) time limit of $H_{max}$, while the specialized model is able to solve $77.8\%$ of the environments. The number of timesteps spent across all environments has lower mean and standard deviation $\mu_{time}, \sigma_{time}$ when the learned prior is used. It is however interesting to note that the specialized model can more easily solve environments that are close to the dynamics it has been trained for. This is why $\min_{time}$, the minimum number of necessary timesteps across all environments is lower for the specialized environment in table \ref{table_specialise}.

\begin{table}[ht]
\begin{tiny}
  \begin{threeparttable}
  \begin{tabularx}{0.98\textwidth} {bbmmmm}
    \ & \ & \ & \\  
    \ & \ & \ & \\
 \hline
     \ \ \ & \#sampled envs & success & $\mu_{time}$ & $\sigma_{time}$ & $\min_{time}$\\
  \hline
    \color{black} general prior \color{black} & \color{black} 600 \color{black} & \color{ForestGreen} \textbf{548 (91.3\%)} \color{black} & \color{ForestGreen} \textbf{121.943} \color{black} & \color{ForestGreen} \textbf{57.85} \color{black} & \color{black} 58 \color{black} \\
  \hline
    \color{black} specialized model \color{black} & \color{black} 600 \color{black} & \color{black} 467 (77.8\%) \color{black} & \color{black} 127.513 \color{black} & \color{black} 92.26 \color{black} & \color{ForestGreen} \textbf{40} \color{black}\\
  \hline
  \end{tabularx}
 \end{threeparttable}
    \caption{\small{Comparison between the prior learned by ACUMEN and a specialized model trained in an average environment on $600$ randomly sampled boxes. Here, $\mu_{time}, \sigma_{time}$ denote the mean and average of the number of timesteps spent in a given environment, which are both lower for the learned prior. The minimum number of time-steps used to solve an environment is denoted $\min_{time}$.}}
\label{table_specialise}
\end{tiny}
\end{table}

\begin{table}[ht]
  \centering
  \includegraphics[width=0.6\linewidth]{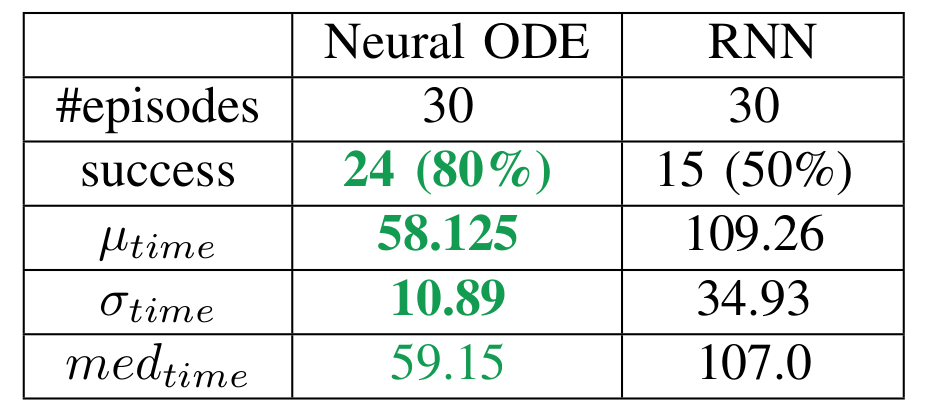}
  \caption{\small{Comparison between a specialized neural ODE and a specialized RNN.}}
  \label{table_rnn}
\end{table}

\begin{figure}[h!]
  \centering
  \captionsetup[subfigure]{justification=centering}
    \subfloat[][]{
      \includegraphics[width=40mm,trim={0cm 0.0cm 0.0cm 0.0cm},clip]{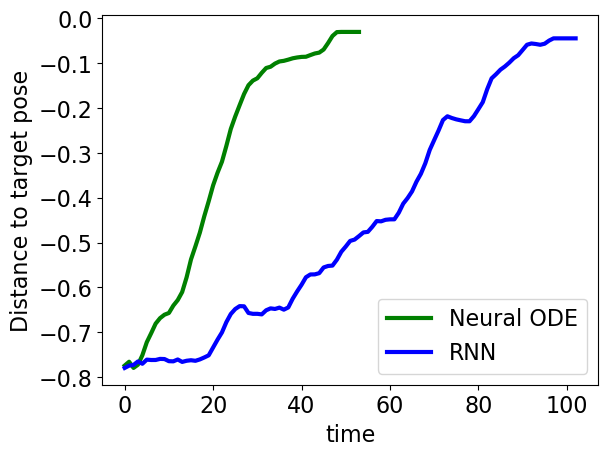}}
    \subfloat[][]{
      \includegraphics[width=39mm,trim={0cm 0.0cm 0.3cm 0.0cm},clip]{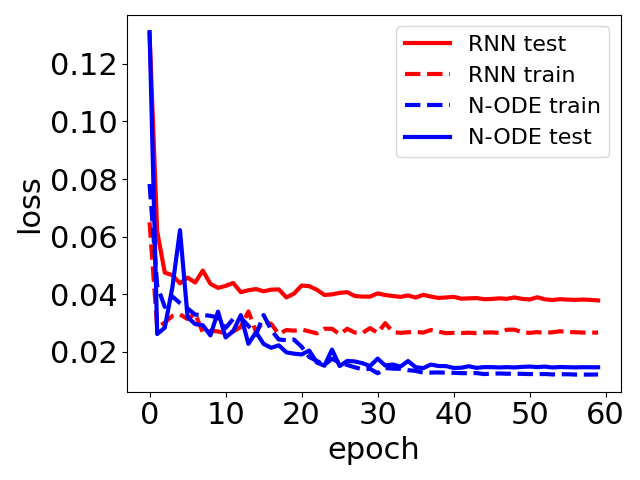}}
    \caption{\small{\textbf{(a)} Comparison between typical trajectories resulting from neural ODEs vs RNNS. \color{black} \textbf{(b)} Train and test losses on the $E_2$ environment.\color{black}}}
  \label{fig_traj}
\end{figure}

\textbf{Neural-ODE ablation.} As discussed in the beginning of section \ref{sec_experim_simu}, we compare the results obtained with a neural ODE with those obtained with a stacked RNN, which takes as input the sequence of tuples $(a_0, \Delta t_0), ... ,(a_n, \Delta t_n)$ with $\Delta t_i$ the elapsed time between $a_i$ and $a_{i+1}$ for $i\neq n$, and $\Delta t_n$ the elapsed time between $a_n$ and the next observation (or during planning, the next predicted state). Note that the RNN and the neural ODE used in this section are comparable in terms of number of parameters ($\sim 9k$ for the RNN vs $\sim 8.5k$ for the neural ODE, see appendix \ref{appendix_hyper} for more details).

For this experiment, we considered an environment with fixed dynamics, that we will note $E_2$, corresponding to a mass of $0.5kg$ and uniform mass distribution. The stacked RNN was first pretrained on data from that environment, which were gathered from three episodes with significant action-state coverage: one with random controls, one successful episode with neural-ODE based control, and another episode using the same neural-ODE but with noise injected into the actions. It was then compared during $30$ model-predictive trials to the specialist neural-ODE model that was learned in the previous section on the $E_1$ environment\footnote{\color{black}We note that both models were trained on approximately the same amount of data, \textit{i.e.} $\sim 400$ observations (with train/test split sizes of $300-400$) and their associated $\sim 1100$ commands.\color{black}}. Note that while the dynamic system is fixed, the observations and actions change from one episode to the other, as they are the result of interpolations at random timestamps, as previously specified in section \S\ref{sec_sim_description}. The results are reported in table \ref{table_rnn}. 

It can be seen that the ODE-based network, specialized for the $E_1$ environment, performs significantly better than the recurrent model on the $E_2$ model, for which the latter is a specialist. Figure \ref{fig_traj}(a) shows two example box pose trajectories from successful episodes, one obtained using the neural ODE (in green), and the other from using the recurrent model (blue curve). Each curve represents the distance from the current pose to the target state. It can be seen that the trajectory obtained with the neural ODE is much smoother than the one resulting from the stacked RNN. 

\color{black} Regarding model accuracy, we also observed that when trained from scratch on the same data from $E_2$ as the RNN, a neural ODE converged to lower loss values on both train/test splits (Figure \ref{fig_traj}(b)). 

We conjecture that the observations above are principally due to 1) the coarser linearization that is implied by the RNN in comparison to the neural ODE and 2) the fact that the RNN has to learn a more complex function. Indeed, in the neural ODE formulation, integration is decoupled from the learned model which only needs to approximate the time derivatives.
\color{black}

\subsection{Gazebo Turtlebot3 simulation}
\label{sec_experim_turtle}

We describe the simulation and dynamics distributions in section \S\ref{sec_gazebo_descr}. In section \S\ref{sec_turtle_ode_vs_rnn}, we define different levels of irregularity in the data, and compare the use of neural ODEs and RNNs for controlling the turtlebot in each of those settings. The complete system's ability to adapt to unseen environment physical properties is then evaluated in section \S\ref{sec_turtle_meta}.

\begin{figure}
\ffigbox[7.8cm]{%
\begin{subfloatrow}
  \ffigbox[\FBwidth][]
    {\caption{}}
    {\includegraphics[width=34mm]{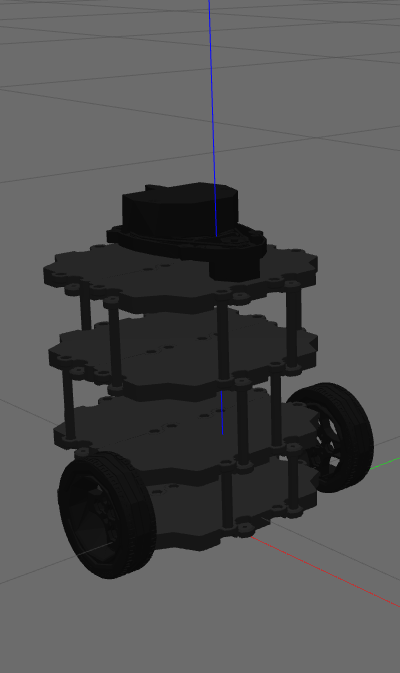}}
\end{subfloatrow}
\begin{subfloatrow}
  \hsize0.7\hsize
  \vbox to 6.35cm{
  \ffigbox[\FBwidth]
    {\caption{}}
    {\includegraphics[width=36mm]{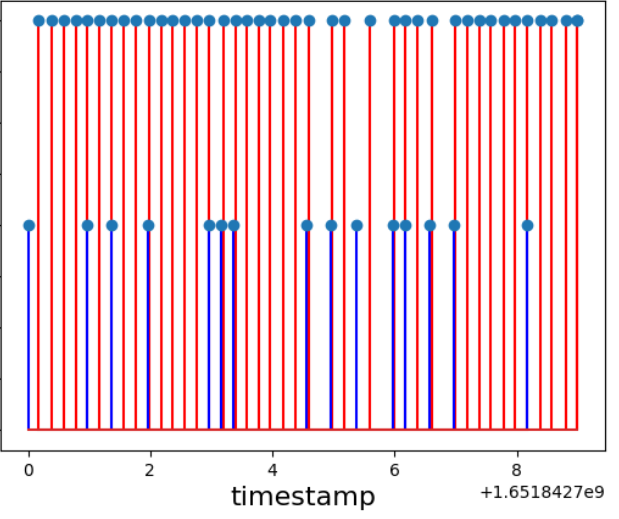}}\vss
  \ffigbox[\FBwidth]
    {\caption{}}
    {\includegraphics[width=36mm]{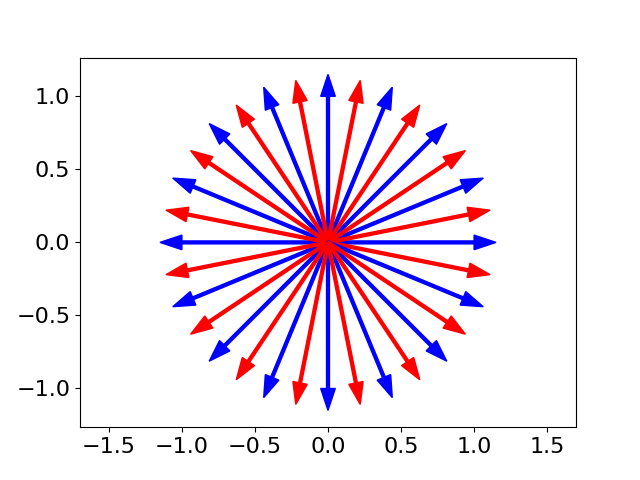}}
  }
\end{subfloatrow}\hspace*{\columnsep}
}
  {\caption{\small{\textbf{(a)} The turtlebot simulation \textbf{(b)} Stem-plot showing an example of observation (in blue) and action (in red) occurrence. In this particular example, observations are dropped with probability $P_{drop}=0.5$, further accentuating the asynchronous/irregular nature of actions and observations. \textbf{(c)} The wind directions used to define the dynamics distributions. Blue arrows indicate directions that are used for training and meta-updates, and red arrows show the directions on which the test distribution is based.}}
   \label{fig_simulation_description_turtle}}
\end{figure}

\begin{figure*}[ht!]
  \centering
      \includegraphics[width=140mm,trim={0cm 0 0cm 0.0cm},clip]{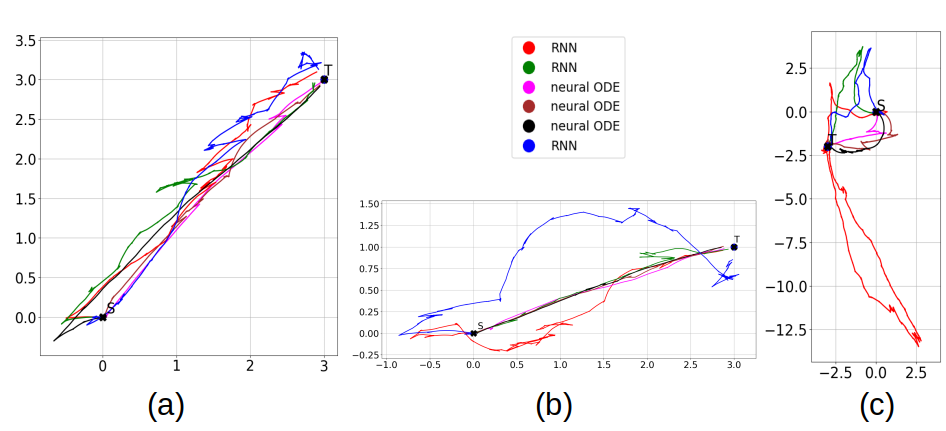}
   \caption{\small{Examples of trajectories obtained with Neural ODEs and RNNs for different values of $P_{drop}$. In all figures, the initial position of the robot and the target pose are respectively noted $S, T$. \textbf{(a)} Trajectories obtained when $P_{drop}=0$. Note that in this case, irregularities in the input data are still present due to the asynchronous nature of ROS. \textbf{(b)} Example trajectories with $P_{drop}=0.2$. \textbf{c} Trajectories obtained with $P_{drop}=0.5$.}}
   \label{fig_turtle_ode_vs_rnn_1}
\end{figure*}

\subsubsection{Gazebo turtlebot3 simulation: description}
\label{sec_gazebo_descr}

This simulation \cite{gazeboturtle} uses the ROS Gazebo \cite{koenig2004design} package and is therefore closer to a real robotic system than the pybullet simulation discussed in the previous section. Furthermore, policies trained in this simulation or more generally with Gazebo simulation can be successfully transferred to real systems, although strategies such as domain randomization are often necessary to avoid overfitting to the simulator \cite{chaffre2020sim, nikdel2021lbgp, zhao2020sim}. A screenshot of the simulation is given in figure \ref{fig_simulation_description_turtle} (a).

The command space of the robot is given by $[-0.22, 0.22] \times [-0.22, 022]$ for linear and angular velocity control \footnote{The maximum absolute value of the angular velocity that can be sent to the turtlebot is $2.84$, but we use a lower value in order to simplify the dynamics.}. We use the pose estimation of the robot \textemdash as published to the \texttt{/Odom} topic by the Gazebo turtlebot3 simulation \textemdash as the observation, but reduce the publication frequency of that information to $\sim 6\text{Hz}$. As in the box rotation experiments, we approximate the state to propagate as $\hat{z}=\xi(\omega_t,\omega_{t-1})\triangleq[\omega_t, \frac{\omega_t-\omega_{t-1}}{\Delta t}]$ where $\omega_{t-1}, \omega_t$ are the two last pose observations.

In all experiments, the robot starts at the origin and its objective is to navigate to an arbitrarily set target. We will first compare the effectiveness of RNNs and neural ODEs when faced with different levels of irregularity, that will be defined by the probability $P_{drop}$ of dropping observations. Note that even $P_{drop}=0$ will result in irregularity in the data as a result of the asynchronous nature of ROS. An example of action and observation occurrences during a randomly selected time interval with value $P_{drop}=0.5$ is shown in figure \ref{fig_simulation_description_turtle}.

The comparisons between neural ODEs and recurrent models which are presented in sections \ref{sec_turtle_ode_vs_rnn} use the default settings of the simulator, where except for gravity, no external forces are applied to the robot. In order to evaluate the ability of the complete system to adapt to unseen environment physics (section \S\ref{sec_turtle_meta}), we additionally define a distribution over physical properties of the simulation which are given by a choice of constant "wind" that is implemented as a constant acceleration applied to the robot. For simplicity, we consider a discrete set of directions, uniformly spaced on the unit circle (figure \ref{fig_simulation_description_turtle}(c)), and a discrete set of magnitudes. Details on the different splits used for training, meta-updates and tests will be given in section \S\ref{sec_turtle_meta}.

Note that the particular instance of algorithm \ref{algo_action_sel} which is used in this section is equivalent to a simple random shooting (appendix \ref{appendix_cmd_sel}). While this sacrifices some precision, it allows the algorithm to publish decisions at a higher rate. The same motivation led us to replace the adaptive-step solver of the previous experiments by a fixed step ODE solver (\texttt{RK4}) which proved sufficient for controlling the turtlebot.

\subsubsection{Gazebo turtlebot3 simulation: neural ODEs vs stacked RNNs}
\label{sec_turtle_ode_vs_rnn}

Here we concentrate on control using an environment with fixed physical properties in order to investigate the advantages of neural ODEs for control in the case of irregular/asynchronous observations. The physical properties are set to the defaults and no wind is applied. As in the previous section, we use a stacked RNN \textemdash the hyperparameters of which are specified in appendix \ref{appendix_hyper} \textemdash as a baseline. Note that both the RNN and the N-ODE are pre-trained until convergence with $P_{drop}=0$ and using continuous actions that are sampled according to a uniform distribution over $[-0.22, 0.22]$ for both the linear and angular velocity controls. However, as detailed in appendix \ref{appendix_cmd_sel}, during control, actions are selected among a discretized set of velocity controls.  

In many real world robotic systems \footnote{We exclude real-time Operating Systems from this discussion.}, different components often publish asynchronously to different topics and at different frequencies based on multiple hardware and software related constraints. This can result in significant irregularities. As controlling the simulated turtlebot is done via those same publication, subscription and service mechanisms, irregularities in actions and observations also naturally occur. However, in these simplified settings, those irregularities are rather small. Therefore, we introduce an additional source of irregularity by dropping each in-coming observation with a probability of $P_{drop}$. In what follows, we compare the RNN and the N-ODE for values of $P_{drop}=0, 0.2, 0.5$. Figure \ref{fig_simulation_description_turtle}(c) shows a stemplot of command/observation occurrences for $P_{drop}=0.5$. 

Figure \ref{fig_turtle_ode_vs_rnn_1} shows a few example trajectories for different values of $P_{drop}$ and arbitrary set target positions. It can be seen that the performance of the RNN drops significantly as $P_{drop}$ increases, while the impact on the neural ODE is limited. Indeed, while the trajectories obtained with the neural ODE remain short and smooth for different $P_{drop}$ values, the trajectories produced when using the stacked RNN become longer and less regular with the increase of $P_{drop}$.

\noindent\textbf{Quantitative results.} We ran a total of $150$ control experiments, in which the initial position of the robot was set to the origin and its orientation to the identity matrix. Each experiment was fully determined by the choice of 
\begin{enumerate}
  \item a pretrained model: RNN or neural ODE,
  \item a target out of five possible positions\footnote{The target position were set over all quadrants of the $xy$ plane in an arbitrarily manner to \{(3.0, -2.0, 0.0), (-3.0, -2.0, 0.0), (3.0, 1.0, 0.0), (3.0, 3.0, 0.0), (2.5, 6.3, 0.0) \}}.
  \item Observation drop probability $P_{drop}$, chosen in the set $\{0.0, 0.2, 0.5\}$.
\end{enumerate}

resulting in $30$ possible combinations, and ran each of them $5$ times. The trajectories were compared according to three characteristics: 

\begin{enumerate}
  \item Trajectory length. This is just the length of the curve.
  \item Irregularity, which we define as $S_{\kappa}\triangleq \int_{a}^{b}\kappa(s)ds$, where $\kappa$ is the unsigned curvature and $ds$ is the length element.
  \item Episode length in terms of number of applied actions.
\end{enumerate}

Indeed, while shorter and smoother trajectories are desirable, a control process that would only chose conservative, small actions might indicate poor predictions from the learned model, hence the consideration of the number of actions as the third characteristic.

\begin{figure}[h!]
  \centering
  \captionsetup[subfigure]{justification=centering}
    \subfloat[][Trajectory length and irregularity for $P_{drop}=0.0$]{
      \includegraphics[width=65mm,trim={0cm 0 0cm 0.0cm},clip]{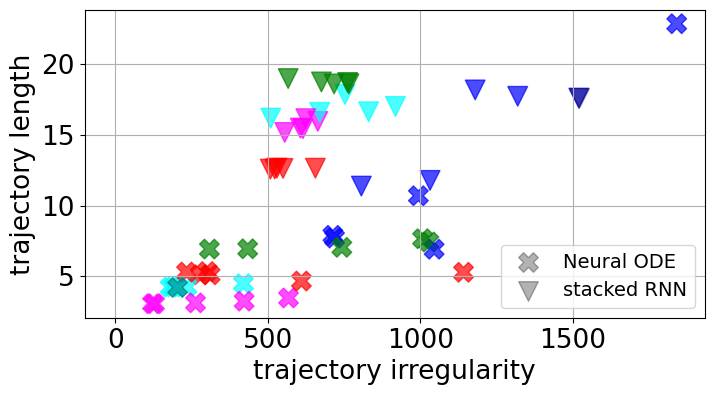}}\\
    \subfloat[][Trajectory length and irregularity for $P_{drop}=0.2$. The dashed lines mark the maximum irregularity and length that were reached in the previous experiment ($P_{drop}=0$, figure \ref{fig_pareto}(a)).]{
      \includegraphics[width=65mm,trim={0cm 0 0cm 0.0cm},clip]{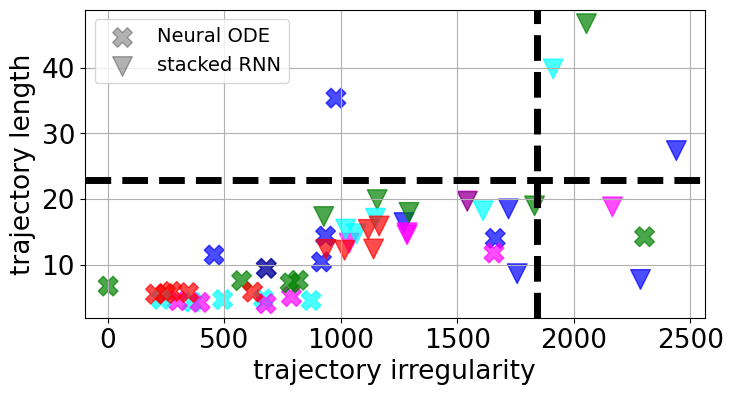}}\\
    \subfloat[][Trajectory length and irregularity for $P_{drop}=0.5$. The dashed lines mark the maximum irregularity and length that were reached in the previous experiment ($P_{drop}=0.2$, figure \ref{fig_pareto}(b)).]{
      \includegraphics[width=65mm,trim={0cm 0 0cm 0.0cm},clip]{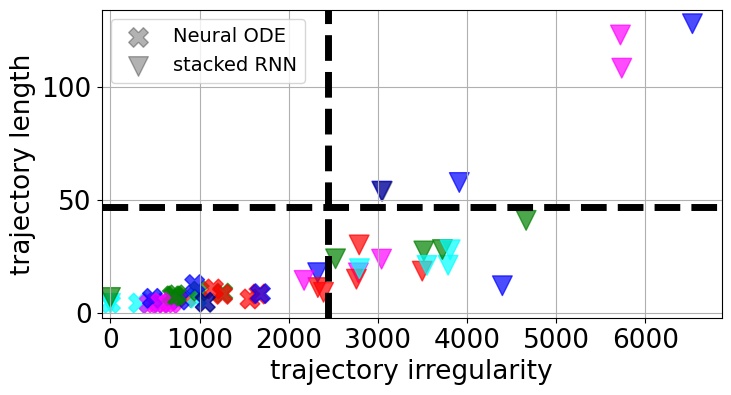}}\\
    \subfloat[Mean and standard deviation of episode lengths (in terms of required actions for success).][Mean and standard deviation of episode lengths (in terms of required actions for success).]{
      \includegraphics[width=55mm,trim={0cm 0 0cm 0.0cm},clip]{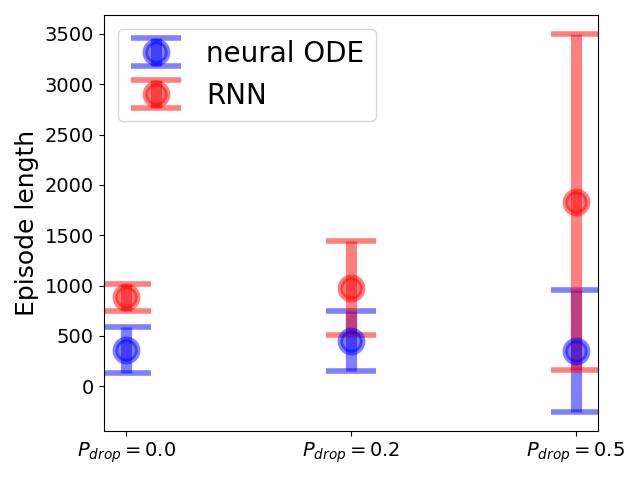}}
    \caption{\small{\textbf{(a), (b), (c)} each correspond to $50$ experiments corresponding to a value of $P_{drop}$, such that half experiments use a neural ODEs and the other half uses a recurrent model. The experiments in which the positional target of the robot was the same are indicated using the same color. \textbf{(d)} compares the length of episodes in terms of number of commands necessary to reach the goal.}}
   \label{fig_pareto}
\end{figure}

The results are reported figure \ref{fig_pareto}. Note that in figures \ref{fig_pareto}(a, b, c), markers of the same color correspond to episodes that had the same positional target for the robot. As it can be seen in figure \ref{fig_pareto}(a), the small irregularities that are present due to the asynchronous nature of ROS in general seem to result in trajectory lengths and irregularities that are slightly larger for the recurrent models. As we increase $P_{drop}$ to $0.2$ (figure \ref{fig_pareto}(b)), the trajectories become longer and less regular for both models. However, the effect is more pronounced for the RNNs, in particular in terms of trajectory irregularity: neural ODEs mostly remain in the same interval ($S_{\kappa}<1000$) as in the previous experiment ($P_{drop}=0$, figure \ref{fig_pareto}(a)), but the majority of RNNs produce trajectories with irregularities $S_{k}>1000$ while that was not the case in the previous experiment with $P_{drop}=0$. As for trajectory lengths, RNNs more frequently produce trajectories with lengths greater than the upper bounds on the results of the previous experiment (dashed lines in figure \ref{fig_pareto}(b)).

The gap between the performances of neural ODEs and RNNs is widened further as $P_{drop}$ is increased to $0.5$ (figure \ref{fig_pareto}(c)). The majority of lengths and irregularities of the trajectories generated by RNNs takes values greater than the upper bounds on previous experiments (dashed line in figure \ref{fig_pareto}(c)). The lengths in particular are increased by factors of up to $5$, while irregularities are increased by factors of up to $3$. On the other hand, length-wise, the trajectories generated by neural ODEs remain in the same interval as in the previous experiments. The increase in irregularity is also much less pronounced than with the recurrent models.

Neural ODEs also outperform the considered RNNs in terms of number of required actions. As shown in figure \ref{fig_pareto}(d), the average number of actions necessary for success using neural ODEs is upper bounded by the corresponding value for RNNs which significantly increases with $P_{drop}$.

\begin{figure}[ht!]
  \centering
  \captionsetup[subfigure]{justification=centering}
    \subfloat[][]{
      \includegraphics[width=40mm,trim={0cm 0 0cm 0.0cm},clip]{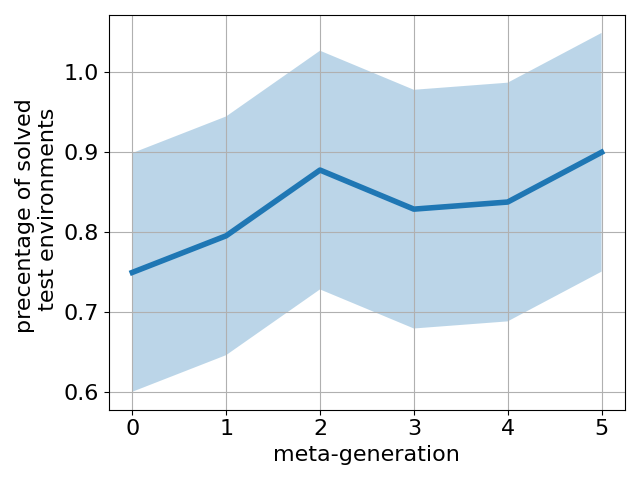}}
    \subfloat[][]{
      \includegraphics[width=40mm,trim={0cm 0 0cm 0.0cm},clip]{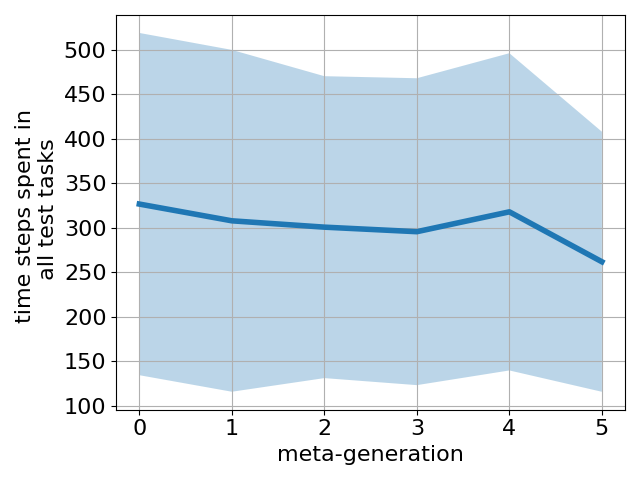}}
    \caption{\small{The effect of a few meta-updates to the pretrained neural ODE, averaged over three distinct execution. Both figures show average values and one single standard deviation.}}
   \label{fig_turtle_meta}
\end{figure}

\subsubsection{Gazebo turtlebot3 simulation: Adaptation to novel dynamics}
\label{sec_turtle_meta}
As mentioned in section \S\ref{sec_gazebo_descr}, a distribution of physical properties based on applying "wind" as a constant acceleration vector defines the dynamics distribution $P(\mathcal{T}_ {turtle})$ that we consider. More formally, noting $S_{mag}\triangleq\{0.1, 0.2, 0.4, 0.5\}$ the set of possible magnitudes, the train and test distributions are given by

\begin{equation}
  \begin{split}
    P(\mathcal{T}_{turtle})^{train}& \triangleq\mathcal{U}(\{[\cos(u), \sin(u)]^T | u=n\frac{2\pi}{n_d}\ \ \ \\ & \forall n=2k\ \ \ ,k\in \mathbb{Z}\}\times S_{mag})\\
    P(\mathcal{T}_{turtle})^{test} & \triangleq \mathcal{U}(\{[\cos(u), \sin(u)]^T | u=n\frac{2\pi}{n_d}\ \ \ \\ &  \forall n=2k+1\  \ \ ,k\in\mathbb{Z}\}\times S_{mag})\\
  \end{split}
\end{equation}

where $\mathcal{U}$ denotes the uniform distribution and $n_d$ is the number of wind directions. In our implementation, we chose $n_d=32$, resulting in $16$ directions for each of the train and test splits (figure \ref{fig_simulation_description_turtle}(c)).

Setting $P_{drop}=0.5$ and starting with the pretrained neural ODE of section \ref{sec_turtle_ode_vs_rnn}, we evaluated the effect of meta-updates, each based on $N=25$ train environments. As the asynchronous nature of the publishing/subscription/service mechanisms can lead to different outcomes on each execution, we averaged the results of three distinct experiments, each making $5$ updates to the initial model. The results, displayed in figure \ref{fig_turtle_meta} (a,b), indicate that just a few iterations result in significant improvements in both the number of solved environments (figure \ref{fig_turtle_meta}(a)) and the number of necessary steps (\ref{fig_turtle_meta}(b)).

\section{Experiments on the SOTO2 robot}
\label{sec_experim_soto}

In this section, we first show the feasibility of controlling an industrial robot using the proposed neural ODE based model predictive approach. \color{black} However, due to limited access to the robot, we present meta-learning results only based on offline logs gathered with robots already operating at customer sites.\color{black}

\subsection{Controlling the robot with neural ODEs}
\label{sec_experim_soto_control}

We focus on the gripper of the SOTO2 robot (figure \ref{intro_figure}) and similar to the simulated experiments of section \ref{sec_experim_simbox}, consider rotating a given box by $90^{\circ}$ as the objective to reach during an episode. A notable difference between this experimental setup and the environment considered in the box rotation simulation of section \ref{sec_experim_simbox} is that the distance between the two conveyor belts can vary from an episode to the next: once the SOTO2 has approached a box, it adjusts the distance between the two conveyor belts according to the box's dimensions before loading it onto them.

The observation and command spaces in this experiment are similar to those of the simulated box rotation experiments (\ref{sec_sim_description}). More precisely, the action space, which corresponds to the independent velocity control of the two conveyor belts is given by $[0.05,0.05]\times [0.05,0.05]$ (in $m/s$). The state is given by $\hat{z}=\xi(\omega_t,\omega_{t-1})\triangleq[\omega_t, \frac{\omega_t-\omega_{t-1}}{\Delta t}]$ with $\omega_t$ denoting the pose of the box at time $t$ \footnote{The planar pose of the box, comprised of $2d$ position and yaw in the plane defined by the conveyor belts, is estimated by a pose-tracking pipeline which takes as input a stream from an RGB-D camera and uses standard computer vision tools. The camera is positioned above at the top of the SOTO2, and is looking down on the gripper.}. As illustrated in figure \ref{fig_soto_distribs} (a), when controlling the gripper, both actions and observations can be applied/received irregularly. 

We pre-trained\footnote{As in the previous experiments, the pretrained model is still refined during each episode, as specified in algorithm \ref{algo_pseudo_main}.} a neural ODE model on nine episodes where and empty box of dimensions $12cm\times 30cm\times 40cm$ (figure \ref{fig_soto_distribs} (b)) was rotated through manual publication of commands, on conveyor belts separated by a distance of roughly $22cms$. This amounts to a total of $891$ recorded commands issued over a time interval of about $5.5min$ and the $1605$ observations that were received during that time. For testing, we considered three different box mass distributions, and paired them with different distances between the conveyor belts, which were fixed at the beginning of each episode. The box mass distributions that were used are shown in figure \ref{fig_soto_distribs} (c, d, e, f). In all experiments, the initial pose of the box up to some variation \footnote{A few centimeters ($<5cms$) in the direction parallel to the conveyor belts, and less than $3cms$ in the direction orthogonal to them.} is similar to the one showed in figures \ref{fig_soto_distribs}(c, e, f). Note that figure \ref{fig_soto_distribs}(d) shows the target position achieved after a successful rotation.

The action selection mechanism of (algorithm \ref{algo_action_sel}) was guided by a dense reward defined using the distance between the current and target poses (appendix \ref{appendix_cmd_sel_soto}). Success and failure were defined in a binary manner: an episode was considered successful if the box reached the target orientation in less than $150$ commands, and its outcome was recorded as failure otherwise. Information on hyper-parameters can be found in appendix \ref{appendix_hyper}.

\textbf{Large box with centered shoe.} In this experiment, a shoe was placed in the center of the large box of size $12cm \times 30cm \times 40cm$ which was used during training. This places the center of gravity near the center of the box, so the mass distribution of the box is relatively similar to that of the empty box that was used for training. A Total of $60$ experiments were performed, such that each group of $10$ successive experiments was associated with a fixed distance between the conveyor belts from the set $C_{dist}\triangleq \{15.08, 16.45, 18.78, 21.06, 23.17, 25.81\}$ (in cms). The results are reported in figure \ref{fig_soto_results} (in blue). As expected, while the solution seems to adapt reasonably well to all different environments, the larger success rates as well as the shortest episode lengths happen when the physical conditions are close to those of the training dataset.

\textbf{Small box with centered shoe.} In this second set of experiments, we used a smaller box of size $12cm \times 20cm \times 30cm$, and again placed a single shoe in it as in figure \ref{fig_soto_distribs} (c). We performed $10$ experiments for each of the conveyor belt distances from $C_{dist}$ that were applicable for this smaller box, \textit{i.e.} we considered distances of $15.08 and 16.45$. The results are reported in figure \ref{fig_soto_results} (in red). While success rate remained at $90\%$, the number of commands needed to reach the target pose was significantly higher, mainly due to the larger number of gradient updates applied to the model. This is not surprising as the difference between the mass distributions of the test environments and the mass distribution the neural ODE was pretrained on is more pronounced than in the previous experiments with the larger box. 

\textbf{Skewed mass distribution.} We also considered a highly skewed mass distribution by placing a weight ($\sim 300gr$) inside a shoe which in turn was placed in on side of the box. We only performed $10$ experiment with a distance of $18.02cms$ between the two conveyor belts. Among those episodes, $70\%$ led to success, and the average and standard deviations of successful episode lengths were respectively of $57.71$ and $32$. Those results are considerably worse than the results obtained with similar distances between conveyor belts in the previous two experiments. Such difficulties in adapting to unseen mass distributions was part of the motivation for the meta-learning components of ACUMEN, which we evaluated in simulation in sections \ref{sec_experim_simbox} and \ref{sec_experim_turtle}.

Before closing this section, we remark that the rewards (appendix \ref{appendix_cmd_sel}) as well as the binary success/failure criterion used in this section do not take additional constraints, necessary for safe operations in industrial environments into account. While the trajectories during each episode were reasonably smooth, a few failures were due to the box getting too close to the sensors at the extreme end of the gripper (seen in the right side of the bottom image of figure \ref{intro_figure}), where the corner of the box was stuck. Such failure cases could have been avoided with more careful reward shaping, hyperparameter tuning, or by a more complex environment model. Furthermore, we observed that during successful experiments with the highly skewed distribution (figure \ref{fig_soto_distribs}(f)), even though the desired box orientation was reached, the final position of the box was asymmetrical with respect to the conveyor belts. One potential solution to that problem would be to endow the agent with the ability to issue an additional command for adjusting the distance between the two conveyor belts. However, this is beyond the scope of this paper.

\begin{figure}[h!]
  \centering
    \subfloat[][]{
      \includegraphics[width=45mm,trim={2.0cm 0.0 0.0cm 0.0cm},clip]{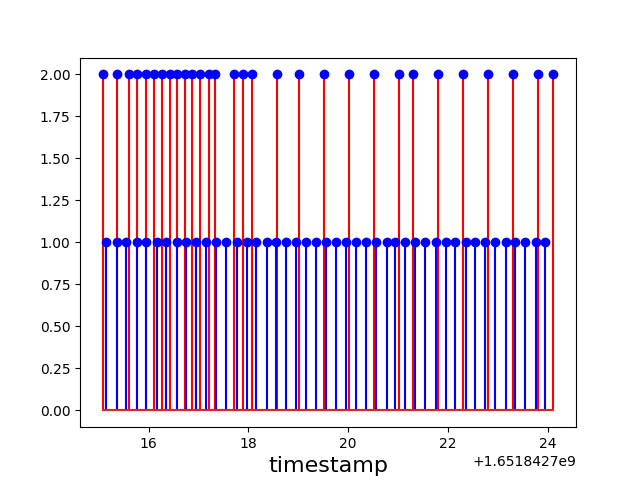}}
    \subfloat[][]{
      \includegraphics[width=40mm,trim={2.0cm 0.0 2.0cm 0.0cm},clip]{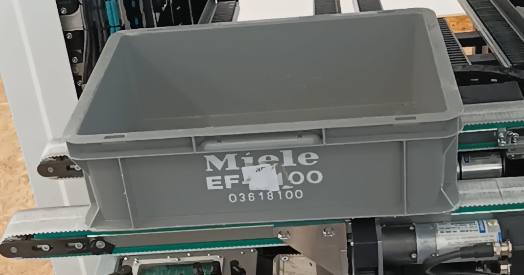}}\\
    \subfloat[][]{
      \includegraphics[width=41mm,trim={0cm 0 0cm 0.0cm},clip]{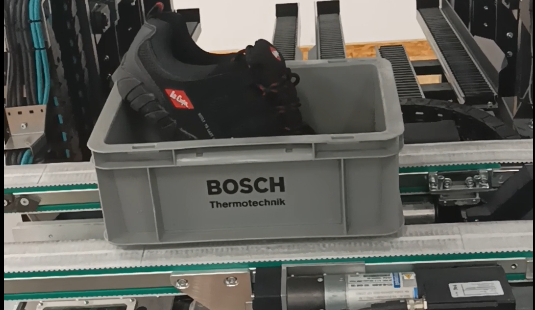}}
    \subfloat[][]{
      \includegraphics[width=43mm,trim={0cm 0 0cm 0.0cm},clip]{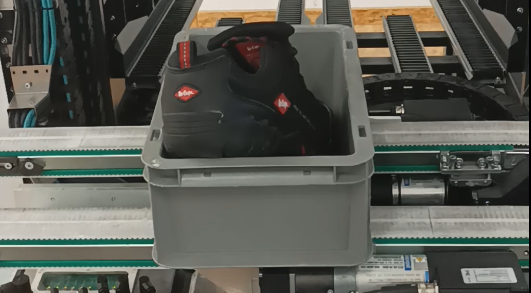}}\\
    \subfloat[][]{
      \includegraphics[width=43mm,trim={1.0cm 0.0 1.2cm 0.0cm},clip]{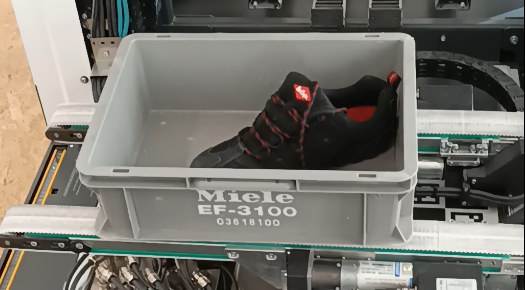}}
    \subfloat[][]{
      \includegraphics[width=42mm,trim={0cm 0 1.0cm 0.0cm},clip]{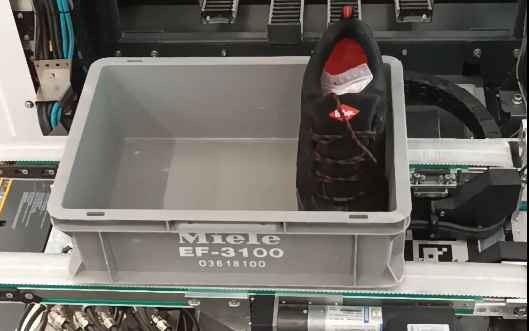}}\\
    \caption{
      \textbf{(a)} Stemplot indicating the occurrence of actions (in red) and observations (in blue) in an experiment with the SOTO2 robot, over an arbitrary time interval. It can be seen that both actions and observations are subject to irregularity.
      \textbf{(b)} The box used for training the neural ODE.
      \textbf{(c, d, e, f)} Different box volumes and mass distributions used for testing. Examples \textbf{(c, d)} indicate the initial state and the target state for the same box, obtained after a successful rotation. While the center of gravity of the boxes shown in \textbf{(c, e)} is close to the center, the distribution displayed in \textbf{(f)} where a small weight is also added to the shoe is highly skewed.}
   \label{fig_soto_distribs} 
\end{figure}

\begin{figure}[h!]
  \centering
    \subfloat[][]{
      \includegraphics[width=41mm,trim={0cm 0 0cm 0.0cm},clip]{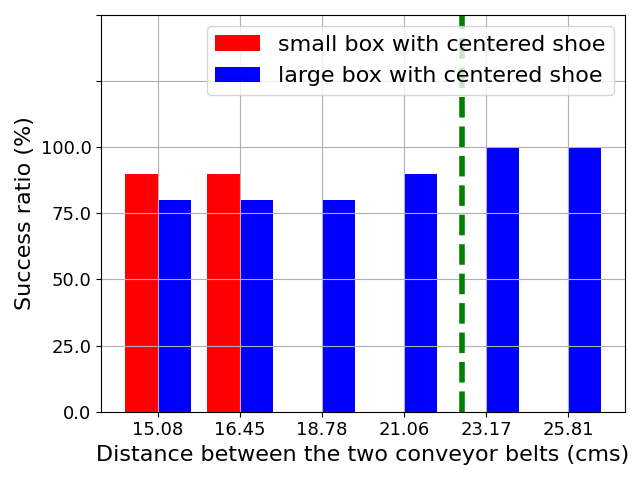}}
    \subfloat[][]{
      \includegraphics[width=43mm,trim={0cm 0 0cm 0.0cm},clip]{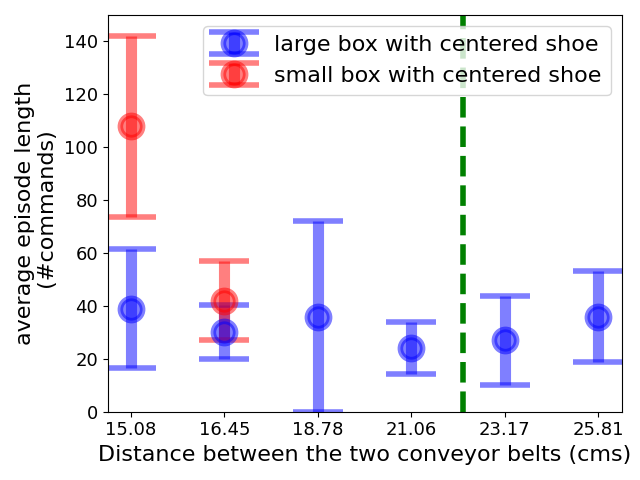}}\\
    \caption{The outcome resulting from using the proposed control method initialized with the pre-trained model. The green dashed line in both figures indicates the approximate distance between the conveyor belts when the training data was collected. \textbf{(a)} The percentage of successful rotations. \textbf{(b)} The average and standard deviation of episodes (in terms of number of commands) for successful rotations.}
   \label{fig_soto_results}
\end{figure}

\subsection{Meta-learning from off-line logs}
\label{sec_experim_soto_meta}

\color{black}
Due to limited access to Magazino's robots, we only present meta-learning experiments based on $1k$ offline logs from robot operating at customer sites gathered while manipulating boxes with various unknown mass distributions. We randomly split this dataset into train, validation and test splits of size $600,200,200$, and meta-learned a model by setting an inner optimization budget of $N_{it}=5$ (\textit{i.e.} the same budget as in the control experiments of the previous section) for each sampled trajectory. The data was \textit{not} collected using our control method, but via a scripted policy which aligns and translates the boxes after they are loaded on the conveyor belts. While this results in lack of diversity in terms of box poses and joint actions that are present in the dataset (figure \ref{fig_meta_logs}(a,b,c)), it should be noted that the boxes handled by the system span a wide array of mass distributions with total masses in the range $[0.1, 30.0]kg$. 

We ran $5$ meta-learning experiments that each had a fixed inner optimization budget of $N_{it}=5$ and a duration of $35$ meta iterations. The averaged results are reported in figure \ref{fig_meta_logs}. Note that as there is little to no rotational motion in the dataset, the orientation approximation error is close to zero and can be ignored. Therefore, figure \ref{fig_meta_logs} only reports translational errors in $cms$. It can be observed that the average error given the budget limit $N_{it}=5$ decreases from about $55mm$ to $46mm$ during the meta optimization. 

\color{black}

\begin{figure}[h!]
  \centering
    \subfloat[][]{
      \includegraphics[width=41mm,trim={0cm 0 0cm 0.0cm},clip]{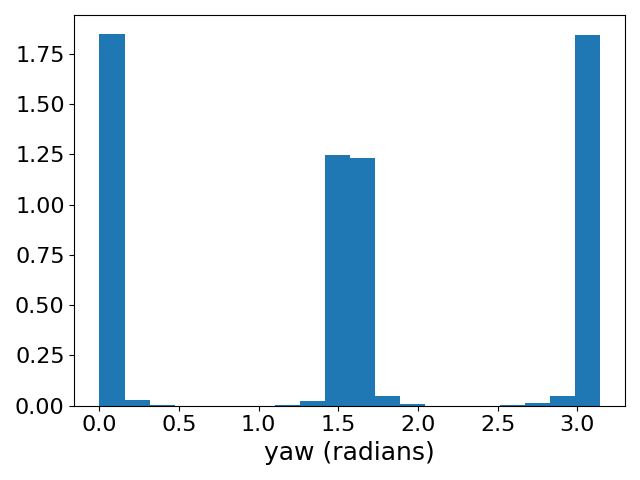}}
    \subfloat[][]{
      \includegraphics[width=41mm,trim={0cm 0 0cm 0.0cm},clip]{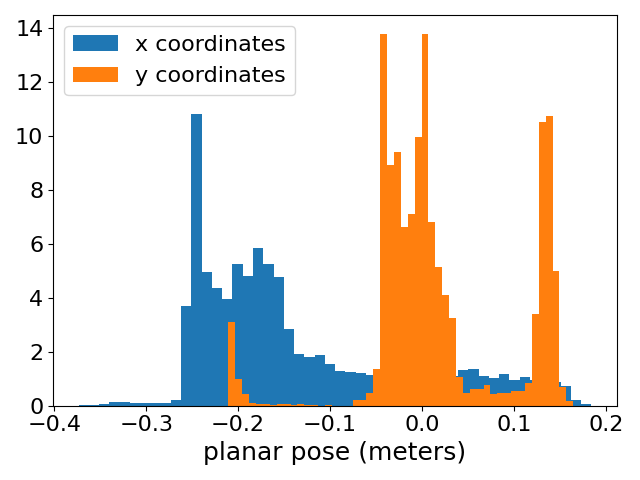}}\\
    \subfloat[][]{
      \includegraphics[width=41mm,trim={0cm 0 0cm 0.0cm},clip]{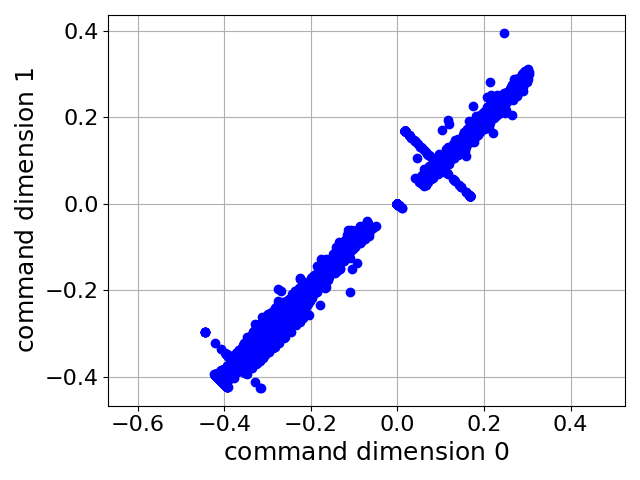}}
    \subfloat[][]{
      \includegraphics[width=41mm,trim={0cm 0 0cm 0.0cm},clip]{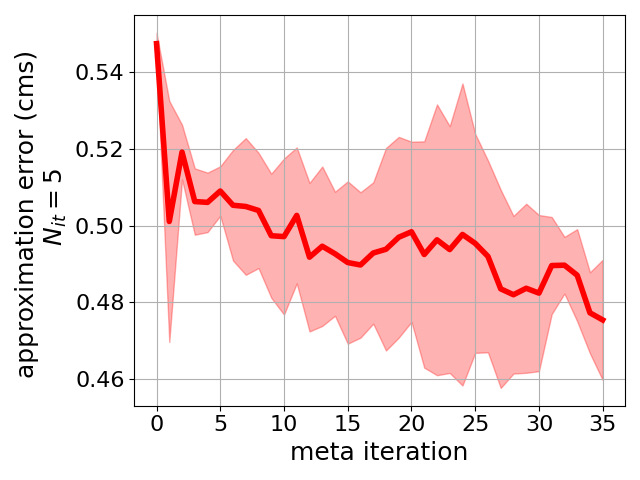}}
    \caption{\color{black}\small{A description of the contents of the offline logs (a,b,c) and the results of meta-learning experiments with $N_{it}=5$. \textbf{(a)} In the logs, the box is either approximately aligned with the conveyors ($\sim 0$ and $\sim 3.14$ radians) or rotated by around $\frac{\pi}{2}$. Therefore, there is almost no rotational motion in that dataset. \textbf{(b)} The planar position mostly varies along the $x$ axis. This corresponds to the translational motion of the scripted policy. \textbf{(c)}. Action dimensions have a correlation of $\sim 1.0$, which also corresponds to the translational behavior of the scripted policy. \textbf{(d)} Average translational error from $5$ meta-learning experiments with $N_{it}=5$, each with a duration of $35$ meta iterations. As there is little to no orientation variation in the dataset, orientation errors, which are close to zero, have not been reported in that plot.}\color{black}}
   \label{fig_meta_logs}
\end{figure}

\begin{figure}[]
  \centering
    \subfloat[\small{A maze navigation task specified in the turtlebot3 gazebo simulation}][\color{black}\small{A maze navigation task specified in the turtlebot3 gazebo simulation}\color{black}]{
      \hspace*{-2.0cm}\includegraphics[width=60mm,trim={0.0cm 0.0 0.0cm 0.0cm},clip]{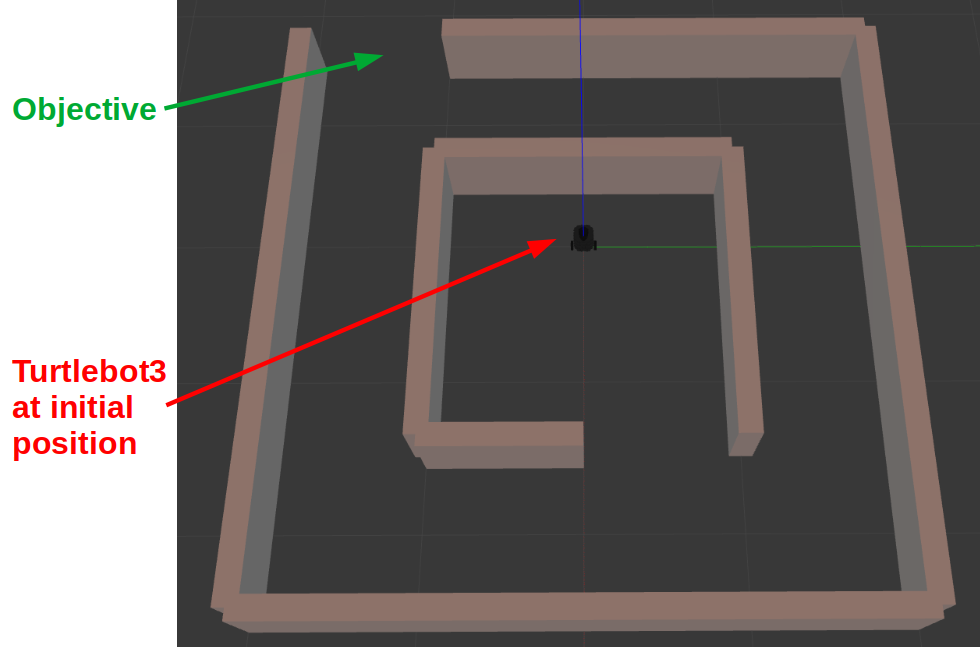}}\\
    \subfloat[\small{Exploration progress using the proposed approach with an exploration-based reward.}][\color{black}\small{Exploration progress using the proposed approach with an exploration-based reward.}\color{black}]{
      \includegraphics[width=81mm,trim={0cm 0 0cm 0.0cm},clip]{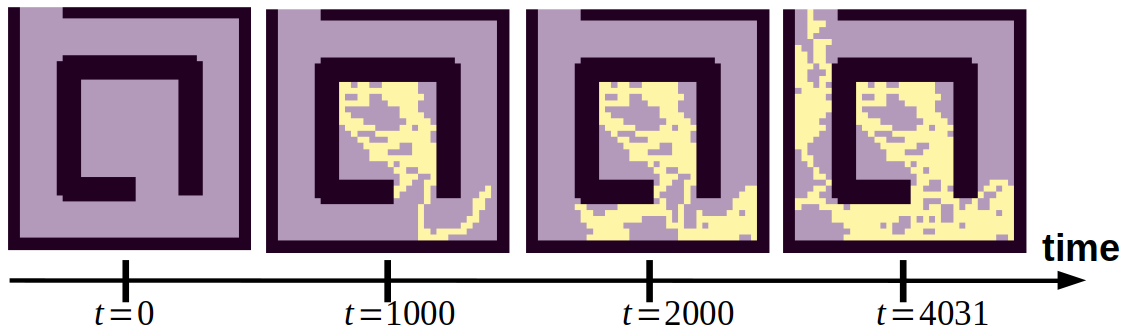}}\\
    \subfloat[\small{Progress of random exploration, which fails to cover enough states to solve the task. This highlights the need for exploration rewards, and thus the advantages of methods, such as ours, that can seamlessly switch between exploration and trajectory-based rewards.}][\color{black}\small{Progress of random exploration, which fails to cover enough states to solve the task. This highlights the need for exploration rewards, and thus the advantages of methods, such as ours, that can seamlessly switch between exploration and trajectory-based rewards.}\color{black}]{
      \includegraphics[width=81mm,trim={0cm 0 0cm 0.0cm},clip]{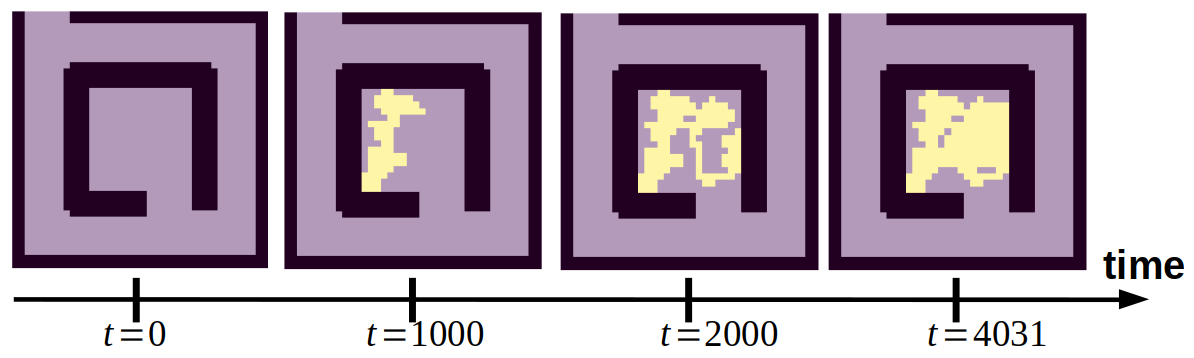}}\\
    \caption{\color{black}\small{A maze navigation task which requires exploration. As the proposed approach is task-agnostic, it can be adapted to new tasks by simply changing the reward function. See appendix \ref{appendix_cmd_sel_exploration} for details on the reward function used for this task.}\color{black}}
   \label{fig_exploration} 
\end{figure}

\section{Discussion}
\label{sec_discussion}

\color{black}
Before discussing the limitations of the proposed approach (\S\ref{sec_limitations}), we highlight some of its advantages compared to MPC approaches that learn the dynamics in a more task-oriented manner.
\subsection{Flexibility}
\label{sec_flexibility}

A desirable property for many robotics systems, especially in open-ended learning settings, is to be able to adapt to both new dynamics and tasks. MPC methods in general trade off flexibility with respect to task changes for accuracy and efficiency in trajectory-tracking in varying dynamics conditions. Indeed, adaptive MPC approaches \cite{richards2022control, richards2021adaptive, spielberg2021neural, jiahao2023online} minimize a function that is dependent on a state-space path (\textit{e.g.} the classical $x^TQx$ term in LQR). However, not all tasks can be specified as trajectories in a straight-forward manner, and it can be preferable to infer such trajectories without explicit planning. This is the case for tasks that require exploration and precise interactions with the environment, including other agents. An example maze navigation task is given in figure \ref{fig_exploration}(a), where the simulated turtlebot3 used in the previous sections is now required to explore a maze. As it can be seen from \ref{fig_exploration}(b), simply swapping the previous goal-based reward function for an exploration reward (appendix \ref{appendix_cmd_sel_exploration}) allows our proposed pipeline to reuse the previously learned neural-ODE model to solve this task, while random exploration (figure \ref{fig_exploration} c) fails. We note that several RL algortihms indeed do achieve similar task-agnostic behavior. However, as discussed in previous sections (\S\ref{sec_intro}, \S\ref{sec_related}), unlike our approach and related MPC methods that leverage neural-ODEs, they do not handle irregular inputs. \color{black}

\subsection{System Limitations}
\label{sec_limitations}

The main limitation of the system lies in its computational complexity (appendix \ref{appendix_computation_time}). While publishing commands at rates of $4-5\text{Hz}$ proved sufficient for controlling the SOTO2 and the turtlebot, we reached those frequencies by switching to a fixed-step solver and by using a simpler action sampling mechanism, sacrificing precision for speed. While this was possible for the SOTO2 and turtlebot experiments, there are many real-world applications were precision and high-frequency control are essential. In order to enable the deployment of the proposed method on such systems, several directions seem worth exploring: \color{black} 1) Parallelism: in our current implementation which has not been optimized, the robot becomes idle during model updates that are made every \textit{\texttt\textrm{train\_freq}} iterations. Parallelising control and model updates, \textit{e.g.} in a manner similar to the one used in Jiaho \textit{et al.}\cite{jiahao2023online} can help alleviate that problem and reduce the resulting latency. 2) Using more efficient solvers/training \cite{poli2020hypersolvers, lehtimaki2022accelerating, kelly2020learning}. 3) Training goal-conditioned policies assuming the availability of a rich joint goal-reward embedding.\color{black}

\section{Conclusion}
\label{sec_concl}

Two challenges are often encountered in industrial robotic systems such as the SOTO2 robot: (i) the asynchronous/irregular publication of observations/actions which violates the assumptions made by the majority of RL algorithms and (ii) dramatic discontinuous changes in environment dynamics from an episode to the next. \color{black} Furthermore, it is desirable for many robotics systems, especially in open-ended learning settings, to be adaptive not only to changes in environment dynamics, but also to changes in tasks. \color{black} Motivated by these observations, we proposed ACUMEN, a task-agnostic Model Predictive Control method that formulates the problem of asynchronous control as a particular case of continuous-time control using learned neural ODEs as environment models, and incorporates meta-learning techniques to ensure adaptivity to changes in dynamics. We evaluated our framework in two simulated environments as well as on the real SOTO2 robot. We also discussed limitations and future directions for improvement. In particular, the computational complexity of the proposed algorithm, while not prohibitive for many use-cases \textemdash such as the studied examples \textemdash remains significant and should be improved in order to make the method applicable to robotic systems requiring high-frequency control.

\section{acknowledgments}This work was supported by the European Union's H2020-EU.1.2.2 Research and Innovation Program through FET Project VeriDream under Grant Agreement Number 951992. We wish to thank Quentin Levent for their initial work on the conveyor belt simulation.  Likewise, we express our gratitude to Patrick Gallinari, Yuan Yin and Kathia Melbouci for their valuable input and criticism during the development of the presented material.

\bibliographystyle{IEEEtran}
\bibliography{example}  

\begin{appendices}
\section{Loss functions}
\label{appendix_loss}
  In order to reduce model complexity and improve stability, as is usual in the literature (\textit{e.g.} \cite{deisenroth2011pilco}), we train the dynamic model to predict changes $\Delta z$ in system state. This process is encapsulated in the function \texttt{OptimizeNODE} from algorithm \ref{algo_pseudo_main}. A wide array of choices can be considered for the loss function, but as our aim is to predict pose in all of our experiments, we use the following weighted least squares formulation:

\begin{equation}
  \mathcal{L}(\mathcal{T}_j, \theta)=\frac{1}{|\mathcal{D}_{train}|}\sum_{b\in \mathcal{D}_{train}}(E^b_{trans}(\theta)+w_lE^b_{rot}(\theta))^2
\end{equation}

  where $E^b_{trans}(\theta)$ and $E^b_{rot}(\theta)$ respectively denote the errors in translation and orientation change predictions over example $b$ from the training data, and where $w_l \in \mathbb{R}$ is a weighting coefficient, used values of which are given in table \ref{table_meta_hyper}. In the presented experiments where displacements are planar, the $E^b_{rot}$ term reduces to an error over the yaw, which in order to avoid discontinuities we simply express as the Frobenius norm of $R_b^T\hat{R}_b$ with $R_b, \hat{R_b}$ respectively the true and predicted rotation matrices.

\section{Command selection}
\label{appendix_cmd_sel}

The reward functions that we have used for each of our experiments are given in the following subsections, where the particular instantiations of algorithm \ref{algo_action_sel} are also discussed.

\subsection{Simulated box rotation}
  \label{appendix_cmd_sel_pybullet}

Let us write $d_{trans}^{init}\triangleq ||z_{trans}^{init}-\hat{z}_{trans}||_2$ the distance between the position predicted by the model and the initial pose of the box on the conveyor belt. Noting $z_{rot}^{target}, \hat{z}_{rot}$ respectively the desired and predicted rotation, the reward for the predicted pose $\hat{z}$ is given by

\begin{equation}
  R(\hat{z})=-||z_{rot}^{target}-\hat{z}_{rot}||_2^2-10.0\mathbb{I}[d_{trans}^{init}> r_t]d_{trans}^{init}
  \label{pybullet_reward}
\end{equation}

  where $\mathbb{I}$ is a predicate function that returns $1.0$ if its argument evaluates to true and returns $0.0$ otherwise. The threshold $r_t$ was set to $0.3$ in our experiments.

\begin{algorithm}
  \SetCommentSty{mycommfontblue}
  \begin{footnotesize}
    \KwIn{Neural ODE weights $\theta$, state approximation $\hat{z}(t_0)$ at time $t_0$, desired duration of state propagation $\Delta t$, previous actions $c_1,...,c_l$ (augmented with their timestamp info), \color{brown}mean and diagonal covariance $\mu_0, \Sigma_0$\color{black}, population size $N_p$, number of elites $N_e$, length of action sequence to sample $H$, reward function $R(.)$, convergence criterion $\mathcal{Y}$, \color{brown}colored noise parameter $\beta$\color{black}}
    \KwOut{action sequence $a^*_1,...a^*_H$} 
    \SetAlgoLined
    \DontPrintSemicolon
    \SetKwFunction{FMain}{CEMStep}
    \SetKwProg{Fn}{Function}{:}{}
    \Fn{\FMain{$\theta$, $\hat{z}(t_0)$,$\{c_1,...,c_l\}$,$\Delta t$, $\mu_0$,$\Sigma_0$,$N_p$,$N_e$,$H$,$R(.)$,$\mathcal{Y}$}}{

      \color{brown}$\mu\leftarrow\mu_0$\\
      $\Sigma\leftarrow\Sigma_0$\color{black}\\
      \tcp{initialize elite set}
      $\texttt{elites}\leftarrow\emptyset$\\
      \While {\texttt{not $\mathcal{Y}$($\Sigma$)}}
      {
        \tcp{sample $N_p$ action sequences of length $H$ at regular intervals in $[t_0, t_0+\Delta_t]$}
        $a^1_{1:H}, ..., a^{N_p}_{1:H}\sim$\texttt{\color{brown}\scriptsize{SampleTimeCorrelatedSequence}\color{black}($\mu$,$\Sigma$,$\beta$)}\\
        \For {\texttt{each} $a^{i}_{1:H}$}
        {
          $\pi_i\leftarrow[c_1,...,c_l]$\texttt{.concatenate(}$a^i_{1:H})$)\\
          \tcp{Let $u_i(t)$ the function that computes actions via interpolating elements of $\pi_i$}
          $u_i\leftarrow\mathcal{I} \circ \pi_i$\\
          \tcp{propagate the state}
          $\hat{z}_i\leftarrow\hat{z}(t_0)+\int_{t_0}^{\Delta t} \hat{\mathcal{F}}(\hat{z}(t),u_i(t),\theta)dt.$\\ 
          \tcp{compute associated reward}
          $r_i\leftarrow$R($\hat{z_i}$)

        }
        $\texttt{elites}\leftarrow$\texttt{select best }$N_e$\texttt{ action sequences according to the }$r_i$\\
        \color{brown}$\mu,\Sigma\leftarrow$\texttt{fit Gaussian to }$\texttt{elites}$\color{black}

      }
      \tcp{return the best action sequence (alternatively, re-sample using $\mu, \Sigma$ after convergence)}
        \KwRet $\texttt{elites}$
    }
    \caption{\small{The particular instantiation of algorithm \ref{algo_action_sel} that is used in the simulated box rotation experiments. It results from sampling time-correlated sequences of actions as in \cite{pinneri2021sample}. Differences with algorithm \ref{algo_action_sel} have been highlighted in brown. The reward $R(.)$ used in this experiment is given by equation \ref{pybullet_reward}.}}
    \label{algo_pseudo_cem}
  \end{footnotesize}
\end{algorithm}

\begin{algorithm}
  \SetCommentSty{mycommfontblue}
  \begin{footnotesize}
    \KwIn{Neural ODE weights $\theta$, state approximation $\hat{z}(t_0)$ at time $t_0$, desired duration of state propagation $\Delta t$, previous actions $c_1,...,c_l$ (augmented with their timestamp info), discrete set of actions ${a_1,...,a_l}$ to sample from, number of actions $N_p$ to sample}
    \KwOut{action sequence $a^*_1,...a^*_H$} 
    \SetAlgoLined
    \DontPrintSemicolon
    \SetKwFunction{FMain}{RandomShooting}
    \SetKwProg{Fn}{Function}{:}{}
    \Fn{\FMain{$\theta$, $\hat{z}(t_0)$,$\{c_1,...,c_l\}$,$R(.)$}}{
      \tcp{sample $N_p$ actions to be applied at $t+\Delta_t$}
        $a_1, ..., a_{N_p}\sim$\texttt{SampleUniform($a_1,..., a_l$)}\\
        \For {\texttt{each} $a_{i}$}
        {
          $\pi_i\leftarrow[c_1,...,c_l]$\texttt{.concatenate(}$a_i)$\\
          \tcp{Let $u_i(t)$ the function that computes actions via interpolating elements of $\pi_i$}
          $u_i\leftarrow\mathcal{I} \circ \pi_i$\\
          \tcp{propagate the state}
          $\hat{z}_i\leftarrow\hat{z}(t_0)+\int_{t_0}^{\Delta t} \hat{\mathcal{F}}(\hat{z}(t),u_i(t),\theta)dt.$\\ 
          \tcp{compute associated reward}
          $r_i\leftarrow$R($\hat{z_i}$)
        }
        $a_{best}\leftarrow$\texttt{select best action according to the }$r_i$\\
        \KwRet $a_{best}$
      }
    \caption{\small{The action selection mechanism used for the turtlebot can be seen as a particular instance of algorithm \ref{algo_action_sel}, which results from setting $H=1$ and choosing a uniform distribution in the latter. The reward function $R(.)$ which guides actions selection in this set of experiments is defined by equation \ref{turtle_reward}.}}
    \label{algo_pseudo_rs}
  \end{footnotesize}
\end{algorithm}

\begin{table*}[ht!]
\begin{tiny}
  \vspace*{0.18cm}
  \begin{threeparttable}
    \begin{tabularx}{0.98\textwidth} {bssmssbbmbms}
 \hline
      \ \ \ & $N$ & $r_{split}$ & $\alpha$ & $\sigma$ & $N_{it}$ & $\tau_{i}.timeout$ & $train_{freq}$ & ODE solver & N-ODE learning rate & $\gamma_{decay}$ & $w_l$\\ 
  \hline
      Simulated box rotation & 20 & 0.75 & 5e-4 & 1e-2 & 10 & 300 & 5 & \texttt{dopri5} & 5e-4 & 0.9 & 10.0\\       
  \hline
      Gazebo Turtlebot3 simulation & 25 & 0.75 & 5e-5 & 5e-3 & 1 & 640 & 40 & \texttt{RK4} & 1e-4 & 0.99 & 1.0\\       
  \hline 
      SOTO2 robot (N-ODE based control)  & N.A & N.A & N.A & N.A & 5 & 150 & 5 & \texttt{RK4} & 1e-3 & 0.99 & 100.0\\       
  \hline 
      \color{black} SOTO2 robot (meta-learning using offline logs) \color{black} & 20 & 0.75 & 5e-2 (decayed) & 1e-3 & 5 & N.A & 5 & \texttt{RK4} & 1e-3 & 0.99 & 100.0\\       
 \end{tabularx}
 \end{threeparttable}
    \caption{\small{Training hyper-parameters for both the outer and lower level optimization problems.}}
 \label{table_meta_hyper}
\end{tiny}
\end{table*}

The actions sampling mechanism that is used in those experiments results from defining the prior distribution $P_{\psi}$ of algorithm \ref{algo_action_sel} as a distribution over time-correlated sequences with noise parameter $\beta$ (\cite{pinneri2021sample}). The complete selection algorithm is given in algorithm \ref{algo_pseudo_cem}.

\subsection{Gazebo Turtlebot3 simulation} 
  \label{appendix_cmd_sel_gazebo}

Keeping the notation for the predicate $\mathbb{I}[.]$ and this time noting $d_{trans}^t \triangleq ||z_{trans}^{target}-\hat{z}_{trans}^t||_2$ the distance between the position predicted by the model and the target position, we write $\hat{z}^t$, $\hat{z}^{t+1}$ the predicted poses at timestamps $t, t+1$. In addition to $d_{trans}$, we consider the difference in angle between the desired heading and the predicted heading:

\begin{equation}
  \begin{split}
    &v_{t}=z_{trans}^{target} - \hat{z}_{trans}^t\\
    &v_{t+1}=z_{trans}^{target} - \hat{z}_{trans}^{t+1}\\
    &u_t^{t+1}=cos^{-1}((\frac{v_t}{||v_t||_2})^T(\frac{v_{t+1}}{||v_{t+1}||_2}))
  \end{split}
\end{equation}

and define the reward as 

\begin{equation}
  R(\hat{z}_{t+1})=-d_{trans}-\mathbb{I}[u_t^{t+1} > r_t^{t+1}]u_t^{t+1}.
  \label{turtle_reward}
\end{equation}

In other words, the error on the heading is ignored when it falls below the threshold $r_t^{t+1}$, which was fixed to $5$° in our experiments.

In contrast with the simulated box rotation experiments, we found that the turtlebot required more frequent changes in controls which could be achieved by setting $\beta=0$ (that is equivalent to sampling according to a Gaussian prior). Furthermore, we found that setting $H=1$ and sampling from a discrete set of velocity values was sufficient to solve the navigation task. A natural additional benefit of this simpler sampling (algorithm \ref{algo_pseudo_rs}) is the reduced computation time required to return a decision.

\subsection{SOTO2 robot} 
\label{appendix_cmd_sel_soto}
  The reward function used on the real robot results from setting $r_t=0.2$ in equation \ref{pybullet_reward}. Regarding command selection, we observed that both cross-entropy based command selection (algorithm \ref{algo_pseudo_cem}) and random shooting (algorithm \ref{algo_pseudo_rs}) were able to lead to successful rotations. However, the latter, when performed over a reduced set of $81$ discretized actions (see appendix \ref{appendix_hyper}) was $\sim 5\times$ faster in terms of execution time. As in the turtlebot experiments, this faster operation time had the additional benefit of allowing faster corrections during box manipulations, leading to an overall lower number of necessary commands to complete the tasks. For these reasons, this sampling method was used throughout the experiments reported in section \ref{sec_experim_soto}.

\subsection{Turtlebot3 exploration task} 
  \label{appendix_cmd_sel_exploration}

  \color{black}  As the aim of this experiment was to highlight the flexibility of the proposed approach rather than optimal navigation, we considered a simple occupancy grid based exploration reward, that we substituted for the previous goal-oriented rewards. We divided the maze environment into $40\times 40$ cells, and recorded the number of times each cell was visited. Then, the reward associated to each state was determined by the number of times the robot had visited the corresponding cell:

  \begin{equation}
    R(\hat{z}_{t})^{\texttt{exploration}}=-\texttt{num\_visits\_to\_cell}(\texttt{cell}(\hat{z}_{t}))
  \end{equation}

  where $\texttt{cell}(.)$ maps each state to the grid cell it falls into. \color{black}

\section{Computation time.}
\label{appendix_computation_time}

  \noindent\textbf{Simulated Box rotation.} We used a standard desktop computer equipped with an AMD Ryzen Threadripper 1920X 12-Core Processor for benchmarking. On average, a single integration of the neural ODE with a batch size of $20$ took $210ms$ while a single forward of the RNN with the same batch size took $3ms$. Similarly, training the neural ODE over a single batch of size $5$ took an average of $380ms$ versus an average of $61ms$ for the recurrent model. Note that the high cost of training and inference with neural ODEs was in part due to the choice of an adaptive-step solver (\texttt{dopri-5}). Furthermore, the sampling-based control used in these experiments (appendix \ref{appendix_cmd_sel}) added some overhead, resulting in decision frequencies of respectively $\sim 10\text{Hz}$ and $\sim 1\text{Hz}$ for the neural ODEs.

  \noindent\textbf{Simulated Turtlebot3 control.} Using the same hardware as in the above, a single forward pass of the RNN with batch size $21$ (one example in the batch per possible discrete action), took on average $4ms$, while the integration for the same batch using the ODE solver (\texttt{RK4} in this case) took on average $82ms$. In other terms, neural ODEs took about $20\times$ more time to produce a decision. Similarly, an epoch of training on $120$ samples took an average of $2.63s$ for neural ODEs and an average of $0.9s$ for RNNs. Those results are hardly surprising as vanilla neural ODEs are notoriously slow \cite{poli2020hypersolvers}. 

That being said, as in this particular application, using neural ODEs can actually lead to lower overall computational costs for higher $P_{drop}$ values. For example, for $P_{drop}=0.5$, the increased length and irregularity of the trajectories produced by the RNNs resulted in execution times that exceeded $35min$ for some episodes, while the maximum length reached for neural ODE based episode was $7.1min$. This is coherent with figure \ref{fig_pareto}(d). 

Note that the complete decision pipeline was able to select actions at about $5\text{Hz}$ when using neural ODEs with discrete action selection.

\noindent\textbf{SOTO2 robot.} The robot was controlled remotely, and the corresponding ros package was executed on a precision 3551 laptop with a Intel(R) Core(TM) i7-10850H @ 2.70GHz CPU. A single prediction with the \texttt{RK4} solver with batch size $81$ (one example in the batch per possible discretized action) took on average $132ms$. The complete decision pipeline was able to publish commands at around $4\text{Hz}$.

\section{Algorithm hyperparameters}
\label{appendix_hyper}

In all experiments, the dynamic model $\hat{F}$ of equation \ref{eq_def} was approximated using feed-forward neural networks with \texttt{tanh} activations, taking as input at each integration time-step $t$ the vector $(t, u(t), \hat{z}_t)^T$ concatenating the time, interpolated command and the propagated state. 

\subsection{Simulations}
 For the simulated box rotation experiments, four hidden layers, each of dimension $48$ were used. For the gazebo turtlebot3 simulation, five hidden layers of dimensions $[64,128,128,64,64]$ were used. The stacked RNNs used in each section had approximately the same capacity  as the neural ODEs they were compared to. While the neural ODE used in the simulated conveyor belts experiments had $\sim 8.5k$ parameters, the corresponding stacked RNN had $5$ hidden layers of dimension $32$, resulting in about $\sim 9k$ parameters. Similarly, the neural ODE used to control the turtlebot had $\sim 37k$ parameters and we used a stacked RNN with $5$ hidden layers of dimension $64$ to match this number of parameters.

  For the simulated box rotation experiments, the CEM-based action selection method (algorithm \ref{algo_pseudo_cem}) with hyperparameters $N_p=20, N_e=5, \mu_0=\textbf{0}, \Sigma_0=I, H=4, \beta=2$ was used. In the case of the turtlebot, we found that sampling commands with no time correlation ($\beta=0$) allowed for quicker recovery from wrong command selections. Furthermore, while the model was pretrained on continuous commands sampled from a uniform distribution over $[-0.22,0.22]$, as mentioned in the previous section (appendix \S\ref{appendix_cmd_sel}, algorithm \ref{algo_pseudo_rs}), we found that choosing the actions from a limited discretized set produced satisfactory results during control: in the presented control experiments, linear and angular velocity control were respectively chosen from $\{-0.1, -0.05, -0.01, 0.0, 0.01, 0.05, 0.1\}$ and $\{-0.1,  0.0, 0.1\}$.

  The hyperparameters for the meta update (line $30$ in algorithm \ref{algo_pseudo_main}) as well as the hyper-parameters for the training of the neural ODEs are reported in table \ref{table_meta_hyper}.

\subsection{SOTO2 robot}

The feed-forward network was comprised of five hidden layers of shapes $[80, 160, 160, 80, 80]$, resulting in approximately $59.7k$ parameters. The empirical results presented in section \ref{sec_experim_soto_control} were produced using the same random shooting procedure as for the turtlebot (algorithm \ref{algo_pseudo_rs}), with action selection over the discrete set $[-0.05, -0.02, -0.01, -0.005,  0.0, 0.005, 0.01, 0.02, 0.05] \times [-0.05, -0.02, -0.01, -0.005,  0.0, 0.005, 0.01, 0.02, 0.05]$. The hyperparameters used for training the neural ODEs are reported in table \ref{table_meta_hyper}. \color{black} Note that the meta learning rate used in section \S\ref{sec_experim_soto_meta} was linearly decayed at each meta iteration by a factor of $0.9$. \color{black}

\end{appendices}

\newpage 
\begin{IEEEbiography}[{\includegraphics[width=1in,height=1.25in,clip,keepaspectratio]{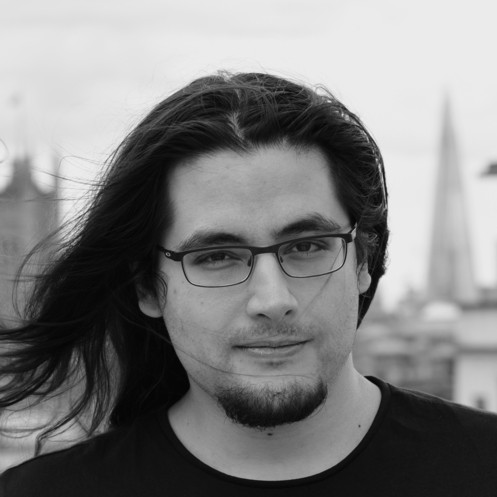}}]{Achkan Salehi}
received the M.S. degree in Computer Science from Sorbonne University in 2014. They conducted their PhD research on SLAM at the French Alternative Energies and Atomic Energy Commission (CEA) and received their PhD from Université Clermont Auvergne in 2018. From 2017 to 2020, they worked on Machine Learning topics related to spatial AI at SLAMCore Ltd. They have since then been with the ISIR lab from Sorbonne University where they primarily work on problems related to generalization and adaptive learning particularly in open-ended contexts.
\end{IEEEbiography}

\begin{IEEEbiography}[{\includegraphics[width=1in,height=1.25in,clip,keepaspectratio]{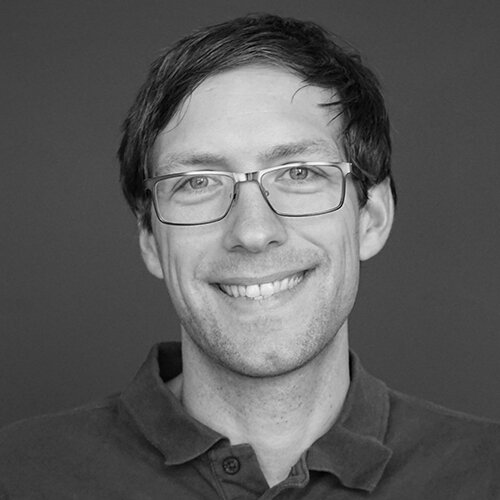}}]{Steffen Rühl}
  received his degree in Computer Science from the University of Karlsruhe in 2007. He received his PhD for his research in the area of planning and execution of manipulation tasks on bimanual robots in variable environments in 2015. The same year he joined Magazino GmbH where today he leads the development of the manipulation process for warehouse robots.
\end{IEEEbiography}

\begin{IEEEbiography}[{\includegraphics[width=1in,height=1.25in,clip,keepaspectratio]{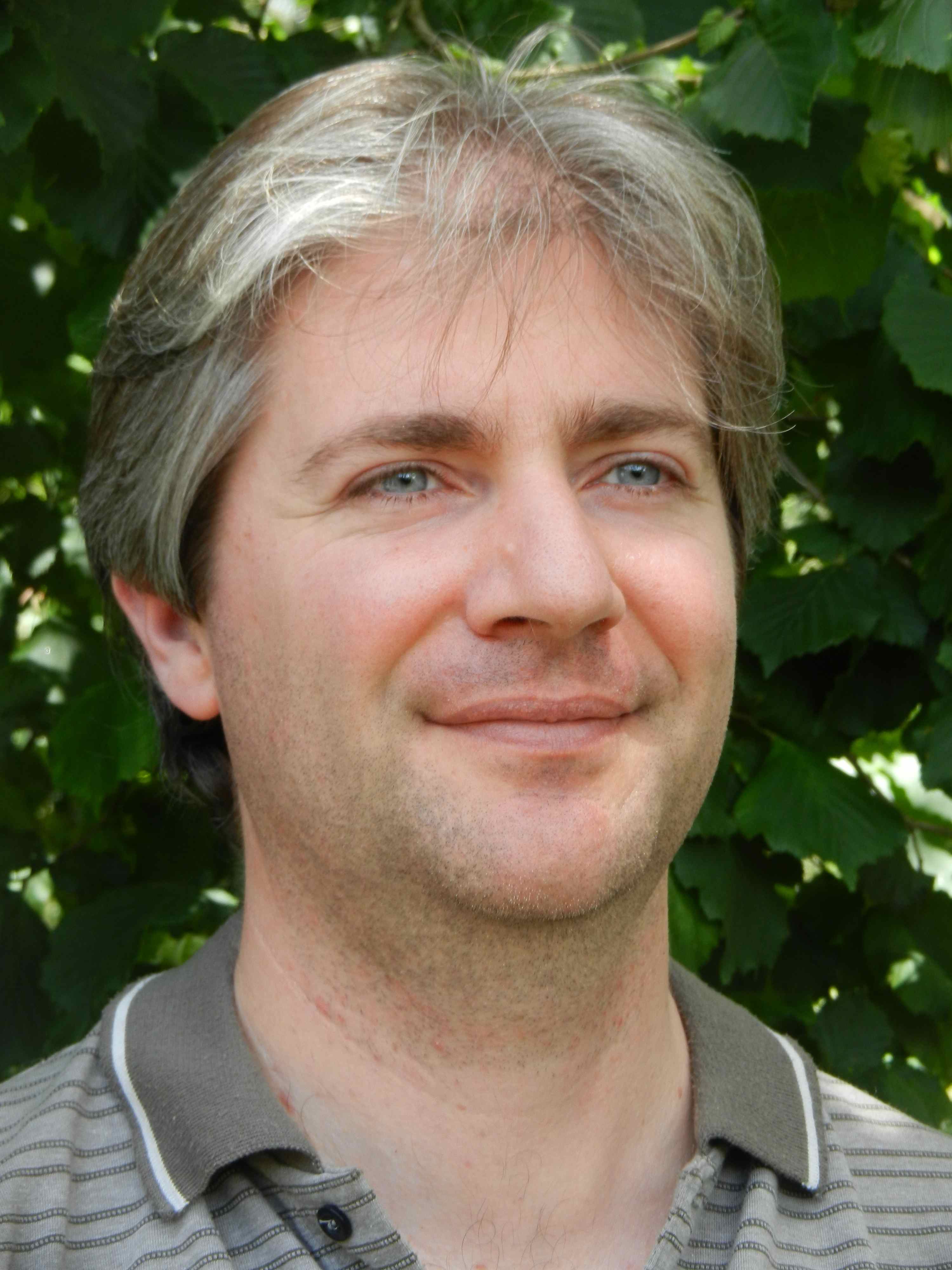}}]{Stephane Doncieux}
 is a Professor of Computer Science at ISIR (Institute of Intelligent Systems and Robotics), Sorbonne University, CNRS, Paris, France. He serves as the deputy director of ISIR, a multidisciplinary robotics laboratory with researchers in automation, mechatronics, signal processing, computer science and cognitive sciences. His research is in cognitive robotics, with a focus on open-ended learning. He focuses on the challenges that robotics raise for learning methods, in particular on the challenge of exploration. Most of his work on this topic is based on Quality Diversity algorithms.
\end{IEEEbiography}

\vfill

\end{small}
\end{document}